\pdfoutput=1

\documentclass[11pt]{article}

\usepackage[final]{acl}

\usepackage{times}
\usepackage{latexsym}

\usepackage[T1]{fontenc}

\usepackage[utf8]{inputenc}

\usepackage{microtype}

\usepackage{inconsolata}

\usepackage{graphicx}
\usepackage{enumitem}
\usepackage{xcolor}
\usepackage{soul}
\sethlcolor{yellow}  
\usepackage{amsmath}
\usepackage{multirow}
\usepackage{booktabs} 
\usepackage{array}
\usepackage{colortbl} 
\usepackage{makecell}
\usepackage{subcaption}
\usepackage{adjustbox}
\usepackage{mathdots}
\usepackage{float}
\usepackage{tcolorbox}
\usepackage{amssymb}
\usepackage[dvipsnames]{xcolor}

\definecolor{darkblue}{RGB}{12, 93, 122}
\definecolor{lightgray}{RGB}{230, 230, 230}
\definecolor{mediumgray}{RGB}{210, 210, 210}

\title{Factuality Beyond Coherence: Evaluating LLM Watermarking Methods for Medical Texts}

\author{Rochana Prih Hastuti, Rian Adam Rajagede, \\ {\bf Mansour Al Ghanim}, {\bf Mengxin Zheng}, {\bf Qian Lou}  \\
        University of Central Florida \\ Orlando, FL \\
        \texttt{\{rochana, rian, mansour.alghanim, mengxin.zheng, qian.lou\}@ucf.edu}}

\usepackage{mdframed}
\usepackage{xcolor}
\newcounter{observation}
\newenvironment{observation}{%
    \refstepcounter{observation}%
    \begin{mdframed}[linecolor=black,linewidth=1pt,backgroundcolor=gray!5]%
    \small
    \textbf{Observation \theobservation:}~\itshape}
    {\end{mdframed}}

\begin{document}

\maketitle

\begin{abstract}

As large language models (LLMs) are adapted to sensitive domains such as medicine, their fluency raises safety risks, particularly regarding provenance and accountability. Watermarking embeds detectable patterns to mitigate these risks, yet its reliability in medical contexts remains untested. Existing benchmarks focus on detection-quality tradeoffs and overlook factual risks. In medical text, watermarking often reweights low-entropy tokens, which are highly predictable and often carry critical medical terminology. Shifting these tokens can cause inaccuracy and hallucinations, risks that prior general-domain benchmarks fail to capture.

We propose a medical-focused evaluation workflow that jointly assesses factual accuracy and coherence. Using GPT-Judger and further human validation, we introduce the Factuality-Weighted Score (FWS), a composite metric prioritizing factual accuracy beyond coherence to guide watermarking deployment in medical domains. Our evaluation shows current watermarking methods substantially compromise medical factuality, with entropy shifts degrading medical entity representation. These findings underscore the need for domain-aware watermarking approaches that preserve the integrity of medical content.

\end{abstract}

\section{Introduction}
LLMs have advanced human-like text generation capability, raising concerns about potentially harmful or biased information in various use case, including in the medical domain \cite{xue-etal-2024-badfair, zhang2025towards,al2023trojbits, loutrojtext, hastuti2025clinic}. Watermarking techniques serve as a safeguard by embedding subtle statistical patterns into generated content that enable detection while preserving quality \cite{kirchenbauer2023watermark, gu2023learnability, kuditipudi2024robust}. 


While watermarking methods show promise for securing information in LLM outputs, their effectiveness in the medical domain remains underexplored \cite{kong2024protecting}. A key limitation is that watermarking often treats all tokens uniformly, ignoring low-entropy tokens \cite{lee-etal-2024-wrote}. These are highly predictable words that often carry critical factual content in medical text, such as disease names. Reweighting them to embed a watermark can alter their meaning, compromising factual accuracy and increasing the risk of hallucinations. Current watermarking evaluation primarily assesses detection-quality tradeoffs from a domain-agnostic perspective, where quality assessment using human or LLM evaluators focuses on measuring text preference \cite{tu-etal-2024-waterbench, molenda-etal-2024-waterjudge} or semantic coherence \cite{singh2024new}, without considering the domain-specific factual risks associated with low-entropy tokens

\begin{figure*}[htbp]
\centering
  \includegraphics[width=0.82\linewidth]{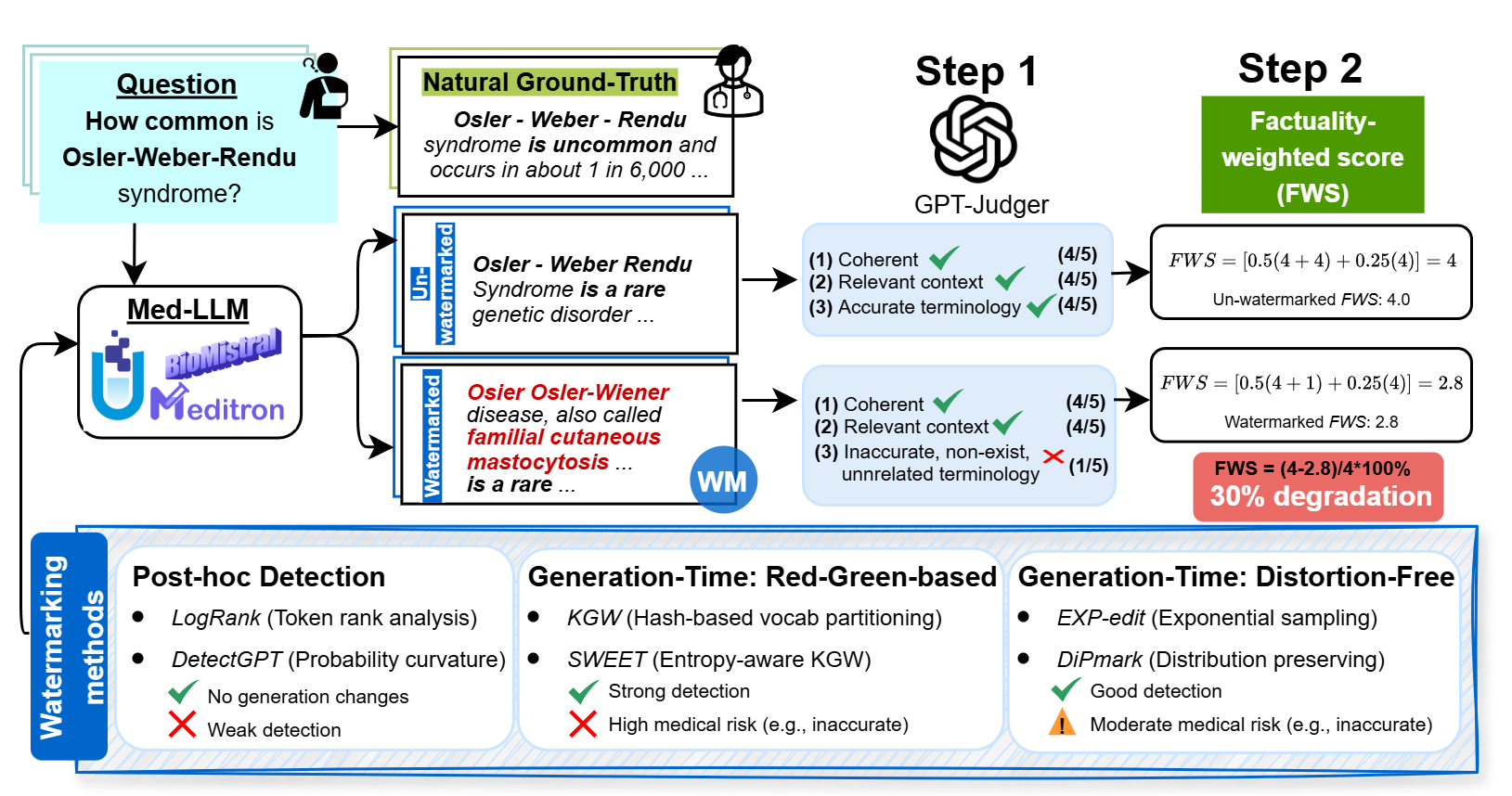} 
  \caption{Factuality degradation in watermarked medical text illustrated through the proposed evaluation framework. \textbf{(1) Evaluation workflow} covers \textit{coherence, relevance,} and \textit{factual accuracy}, applicable to GPT-Judger and traditional metrics. \textbf{(2) Factuality-weighted Score (FWS)} emphasizes critical factual accuracy beyond coherence and serving as a unified metric to guide watermarking deployment in medical applications.}
  \label{fig:benchmark-pipeline}
\end{figure*}

To address the limitations of general evaluation workflows, our watermarking evaluation for medical text is illustrated in Figure~\ref{fig:benchmark-pipeline}. (1) \textbf{Evaluation Workflow}: a unified framework to accommodate both existing automatic metrics and GPT-Judger assessment while examining the critical dimensions of factuality and coherence. This workflow is deliberately flexible, allowing traditional metrics to be integrated alongside our fine-grained GPT-Judger assessments. The key innovation is in creating a structured evaluation that works across different measurement approaches and tasks. (2) \textbf{Factuality-Weighted Score (FWS)}: To reflect the critical nature of factual integrity in medical applications, we introduce a composite quality metric that emphasizes factual correctness beyond coherence. We validate this approach through human evaluation, confirming alignment between expert judgments, our GPT-Judger assessments, and the proposed FWS metric, creating a reliable workflow for evaluating watermarked medical text.

Our analysis reveals critical trade-offs in watermarking methods for medical text. We find that generation-time watermarks, where watermarks are embedded during the LLM’s text generation process, achieve high detection rates but alter token entropy distributions, particularly affecting medical entities. Our evaluation workflow demonstrates that factuality degradation is often more severe than coherence loss, a distinction missed by standard quality metrics. Human evaluations show stronger agreement with our FWS. These results highlight the need for tailored watermarking approaches in medical domains where factual integrity is crucial.

We recommend future approaches: (1) develop domain-aware techniques that preserve low-entropy tokens of medical terminology and avoid hallucination while maintaining detectability, and (2) perform factuality-focused evaluation beyond general quality assessments. These insights aim to develop reliable watermarking methods balancing detectability with crucial medical factuality. 

To summarize, our contributions are four-fold: 
\begin{itemize}
    \item We introduce a unified evaluation framework for medical texts that integrates coherence and factuality dimensions, compatible with both traditional metrics and our proposed GPT-Judger approach.

    \item We develop the Factuality-Weighted Score (FWS) that emphasizes factuality in medical contexts, validated through correlation analysis with human evaluations involving medical practitioners.
    
    \item We provide baseline results on medical tasks, showing that despite high detection capabilities, current watermarking methods face notable factuality degradation.
    
    \item We identify entropy distribution shifts in medical entities as a contributing factor behind factuality degradation, providing critical insights for developing a domain-aware watermarking approach.

\end{itemize}

The code and experimental data are available at: \url{https://github.com/rochanaph/fact-eval-wllm}

\section{Related Work}
\noindent\textbf{Watermarking methods for LLMs}.\hspace{0.5em} Methods to detect LLM-generated text has been studied in previous works, like leveraging the statistical properties without changing the text itself \cite{mitchell2023detectgpt, gehrmann2019gltr}, adding lexical feature as a pattern to distinguish \cite{he2022protecting}, as well as training both traditional and neural-based classifier to classify human-generated and model-generated texts \cite{openai, guo2023close, singh2024new}. Recently, a series of work embeds watermarks within LLM-generated texts by either modifying logits from the LLM \cite{kirchenbauer2023watermark, kirchenbauer2024reliability}, adding a selective technique tailored for domain requirements \cite{lee-etal-2024-wrote}, manipulating the sampling procedure \cite{christ2024undetectable, kuditipudi2024robust}, or investigate the learnability of watermarks in the distillation process \cite{gu2023learnability}.

\noindent\textbf{Benchmarking watermarking methods}.\hspace{0.5em} Existing benchmarking for watermarking methods has defined evaluation criteria, particularly in general-domain scenarios. Most benchmarks emphasize detectability to ensure watermark robustness \cite{tu-etal-2024-waterbench} and complement this with quality assessments often based on LLM-as-evaluator preferences \cite{singh2024new}. However, these evaluations typically lack domain-specific considerations and rely heavily on automatic metrics \cite{ajith2024downstream} or simplified pairwise comparisons, with no further human validation or detailed justification of results \cite{molenda-etal-2024-waterjudge}.

\noindent\textbf{Securing medical LLM's}. The development of medical language models supports the applicability of various tasks in broader clinical and healthcare settings. Despite this progress, efforts to incorporate safeguards for medical language models remain limited, such as in the jailbreaking case or model watermarking via backdoor \cite{kong2024protecting, xue2023trojllm, ghanim-etal-2024-jailbreaking, hastuti2025clinic,zheng2024trojfsp}. Given the relatively recent development of LLM watermarking techniques, their applicability and effectiveness in critical domains like medical remain unexplored and need further investigation.


\section{Evaluation Workflow}
\textbf{General workflow}\hspace{0.5em} 
Our framework is illustrated in Figure~\ref{fig:benchmark-pipeline}.  First, we feed the Medical LLMs with a prompt such as a medical question, and then we evaluate the quality of its watermarked and unwatermarked output. We evaluate watermarking methods across three key quality aspects.
\textbf{1) Coherence} captures the fluency of the generated text and is commonly used in watermarking evaluation. 
\textbf{2) Relevance/Completeness} measures whether the context surrounding critical medical terminology is preserved, ensuring watermarking mechanisms do not distort model confidence.\footnote{Variations in symptom descriptions may correspond to distinct diagnoses, a model should not fluctuate in confidence when generating crucial terms like \textit{"cancer"}. See Figure ~\ref{fig:plot-boxplot_entity}.} 
\textbf{3) Factual accuracy} assesses whether newly introduced medical terminology preserves correct semantics to avoid hallucination \footnote{Entity errors source to factuality hallucination \cite{li-etal-2024-dawn}. See Table \ref{tab:entity-hallucination} for rate and Table \ref{tab:halu} for example.}. The latter two aspects emphasize potential factual corruption, with detailed analysis provided in Section~\ref{subsec:rq4}.

In \textbf{Step 1}, we implement and evaluate these three quality aspects through two approaches:
\begin{itemize}
    \item \textbf{GPT-judger evaluation.} We employ GPT-4o-2024-08-06 as an LLM evaluator to provide fine-grained judgments along the dimensions defined in Table~\ref{tab:quality-aspects}. 
    
    \item \textbf{Traditional automatic metrics.} Our framework also integrates established metrics to enable unified quality quantification. Coherence is measured using non-reference metrics such as Perplexity \cite{kirchenbauer2023watermark} or reference-based similarity measures such as SimCSE \cite{huo2024token}. Accuracy is evaluated with task-oriented, reference-based metrics including \verb|ROUGE-2|, \verb|ROUGE-L|, \verb|F1| \cite{he2024can}, and factuality metric \verb|AlignScore| \cite{zha2023alignscore}, using natural text as the ground truth.
\end{itemize}

In \textbf{Step 2}, we evaluate the overall quality by proposing a quality metric \textbf{Factuality-weighted Score} to see how the watermarking method affects factuality.

\noindent\textbf{Factuality-weighted Score (FWS)}\hspace{0.5em} 
The proposed aspects allow us to quantify the overall quality metric as shown in Equation~\ref{eq:factuality-weighted}.  To prioritize relevant context and factual accuracy over coherence, we set $\alpha=0.4$ and $\beta=0.2$ to indicate that Relevance or Completeness ($Rel.$) and Accuracy ($FactAcc.$) are equally important, and they are twice as important as Coherence ($Coh.$). This weighting was based on empirical consideration through sensitivity analysis provided in Table \ref{tab:config-correlation}.
\begin{equation}
\label{eq:factuality-weighted}
FWS = \alpha(\text{Rel} + \text{FactAcc}) + \beta(\text{Coh})
\end{equation}
FWS serves as a unified metric that enables actionable deployment decisions for watermarking methods in the medical domain and is also applicable to other high-stakes domains where factual accuracy is critical.\\

\begin{table*}[t]
\small
\begin{tabular}{|p{3.7cm}|p{5.5cm}|p{5.3cm}|}
\hline
 \textbf{Text Completion} & \textbf{Question Answering} & \textbf{Summarization} \\
\hline

 \multicolumn{3}{|p{12.5cm}|}{\textbf{Coherence:} Whether the language is coherent, clear, and understandable.} \\
\hline

\textbf{Relevance:} Whether the text includes \textbf{relevant information} for the prompt. & 
\textbf{Relevance:} Whether the answer addresses the question \textbf{without going off-topic and covering all essential parts} & 
\textbf{Completeness:} Whether the generated summary \textbf{misses any important information} from the original text. \\
\hline

\multicolumn{3}{|p{15.5cm}|}{\textbf{Factual Accuracy:} Whether the generated text \textbf{introduces inaccurate or unrelated medical terms not present} in the original reference.} \\
\hline
\end{tabular}
\caption{Proposed quality aspects for evaluating watermarked medical text include \textit{coherence $\mathbf{(Coh.)}$}, \textit{relevance $\mathbf{(Rel.)}$}, and \textit{factual accuracy $\mathbf{(FactAcc.)}$} across text completion, question answering, and summarization. $\mathbf{Rel.}$ assesses contextual information retention, and $\mathbf{FactAcc.}$ ensures the semantics of medical terminology are preserved, both emphasized in the FWS metric as critical for medical applications.}
\label{tab:quality-aspects}
\end{table*}

\noindent\textbf{Human evaluation}\hspace{0.5em} To evaluate both automatic metrics and GPT-Judger ratings on equal footing, we conducted a human evaluation with 6 respondents, including graduate students and medical practitioners, ensuring familiarity with LLMs and medical expertise. The evaluation focused on the KGW watermarking method applied to the Question Answering (QA) task, chosen as KGW underlies many LLM watermarking approaches and QA represents the most realistic practical use case for medical language models. Each item was rated by 3 respondents to obtain an average score using the quality aspects in Table~\ref{tab:quality-aspects}. The full evaluation template is provided in Appendix~\ref{sec:appendix_human}.

Pearson correlation was computed to measure agreement between human judgments and both traditional automatic metrics and GPT-Judger ratings. Prior to this, a Nemenyi test was applied to detect significant differences among the proposed aspects, with lower p-values indicating greater distinction \cite{fu2024qgeval}.

\section{Experiments}
\label{sec:experiments}

\subsection{Research Questions}
\label{subsec:rqs}

Our experiments aim to answer the following research questions:

{\small
\begin{enumerate}[label=\textbf{RQ\arabic*}, leftmargin=1.2cm, rightmargin=0.5cm,itemsep=1pt, parsep=0pt]
    \item How do watermarking methods perform in \textit{detection} and \textit{coherence quality}?
    \item How do watermarking methods perform on \textit{task-oriented} evaluations?
    \item How does the proposed \textit{factuality-weighted score} perform, and is it aligned with human evaluation?
    \item What are the risks and contributing factors of medical \textit{factuality corruption}?
\end{enumerate}
}

\subsection{Watermarking Methods}
\label{subsec:methods}

We evaluate six watermarking methods across two categories \cite{liu2024survey}. \textbf{Post-hoc watermarking} applies after text generation. We evaluate two schemes from this category: LogRank \cite{gehrmann2019gltr}, which classifies text by thresholding the average log-rank of tokens, as machine-generated text typically has smaller average log-ranks, and DetectGPT \cite{mitchell2023detectgpt}, which exploits the tendency of model-generated text to occupy negative curvature regions in the log probability function by comparing original passages with semantically similar perturbations.

The second category is \textbf{generation-time watermarking}, where watermarks are embedded during the LLM's text generation process. We evaluate four approaches: \textit{KGW} \cite{kirchenbauer2023watermark}, a foundational logit-based method that partitions the vocabulary into red and green lists based on a previous token hash; \textit{SWEET}  \cite{lee-etal-2024-wrote}, an improvement of the KGW method on addressing low entropy tokens for code generation domain However, another work \cite{huo2024token} showed that SWEET works well in the general domain, thus we believe potentially useful in medical text where low entropy tokens can be found; \textit{DiPmark} \cite{wu2024resilient} a distribution-preserving watermark that uses complementary reweighting strategies with random token permutations to maintain original text quality; and \textit{EXP-edit} \cite{kuditipudi2024robust}, a token sampling-based method modifying probabilities based on token rank. 

This selection allows comparison across three strategies \textbf{\textit{post-hoc} (PH), \textit{logit-based} (LB)}, distribution preserving or \textbf{\textit{distortion-free} (DF)}, while also includes methods addressing potential domain-specific challenges like low entropy. All generation-time watermarks used the MarkLLM toolkit \cite{pan-etal-2024-markllm} with default parameters. Brief methods are available in Appendix \ref{sec:appendix_method}, and parameter details are available in Appendix \ref{sec:appendix_hyperparameter}, Table \ref{tab:defaultparam}.

\begin{table*}[h]
\centering
\tiny
\resizebox{\textwidth}{!}{
\begin{tabular}{clcccc|cccc|cccc}
\hline
\multirow{3}{*}{\footnotesize\textbf{Schm.}}& \multirow{3}{*}{\small \textbf{\makecell[l]{Waterm. \\Methods}}} & \multicolumn{4}{c|}{\small \textbf{Meditron 7B}} & \multicolumn{4}{c|}{\small \textbf{BioMistral 7B}} & \multicolumn{4}{c}{\small \textbf{MedLlama 8B}} \\
& & \multicolumn{2}{c}{\scriptsize \textbf{Detection}} & \multicolumn{2}{c|}{\scriptsize \textbf{Quality}} & \multicolumn{2}{c}{\scriptsize \textbf{Detection}} & \multicolumn{2}{c|}{\scriptsize \textbf{Quality}} & \multicolumn{2}{c}{\scriptsize \textbf{Detection}} & \multicolumn{2}{c}{\scriptsize \textbf{Quality}} \\
\arrayrulecolor{gray!40}\cline{3-14}\arrayrulecolor{black}
& & \scriptsize \textbf{TPR}$\uparrow$ & \scriptsize \textbf{AUROC}$\uparrow$ & \scriptsize \textbf{PPL}$\downarrow$ & \scriptsize \textbf{SimCSE}$\uparrow$ & \scriptsize \textbf{TPR}$\uparrow$ & \scriptsize \textbf{AUROC}$\uparrow$ & \scriptsize \textbf{PPL}$\downarrow$ & \scriptsize \textbf{SimCSE}$\uparrow$ & \scriptsize \textbf{TPR}$\uparrow$ & \scriptsize \textbf{AUROC}$\uparrow$ & \scriptsize \textbf{PPL}$\downarrow$ & \scriptsize \textbf{SimCSE}$\uparrow$ \\
\hline
&\multicolumn{13}{>{\columncolor{blue!20}}c}{\scriptsize\textit{Text Completion}} \\
\rowcolor{gray!20} \cellcolor{white} &  Un-watermarked
& -- & -- & 9.7 & 1.000
& -- & -- & 9.2 & 1.000
& -- & -- & 4.5 & 1.000 \\
\multirow{2}{*}{PH} & LogRank
& 0.063 & 0.764 & 9.7 & 1.000
& 0.146 & 0.750 & 9.2 & 1.000
& 0.042 & 0.700 & 4.5 & 1.000 \\
& DetectGPT
& 0.010 & 0.704 & 9.7 & 1.000
& 0.021 & 0.592 & 9.2 & 1.000
& 0.010 & 0.619 & 4.5 & 1.000 \\
\arrayrulecolor{gray!40}\cline{3-14}\arrayrulecolor{black}
\multirow{2}{*}{LB} & KGW
& 0.999 & 0.999 & 12.6 & \textbf{0.645}
& \textbf{1} & \textbf{1} & 11.6 & 0.688
& \textbf{0.890} & \textbf{0.995} & 4.9 & \textbf{0.782} \\
& SWEET
& \textbf{1} & \textbf{1} & 12.7 & 0.642
& \textbf{1} & \textbf{1} & 11.4 & 0.685
& 0.860 & 0.994 & 4.6 & 0.765 \\
\arrayrulecolor{gray!40}\cline{3-14}\arrayrulecolor{black}
\multirow{2}{*}{DF} & DiPmark
& 0.990 & 0.999  & \textbf{10.9} & 0.643
& \textbf{1} & \textbf{1} & \textbf{9.9} & \textbf{0.709}
& 0.290 & 0.944 & \textbf{4.5} & 0.776 \\
& EXP-edit
& 0.935 & 0.991 & 26.6 & 0.625
& 0.980 & 0.996 & 12.8 & 0.654
& 0.040 & 0.689 & 9.8 & 0.758 \\
\cline{2-14}
&\multicolumn{13}{>{\columncolor{blue!20}}c}{\scriptsize\textit{Question Answering}} \\
\rowcolor{gray!20} \cellcolor{white} & Un-watermarked
& -- & -- & 8.7 & 1.000
& -- & -- & 8.5 & 1.000
& -- & -- & 4.2 & 1.000\\
\multirow{2}{*}{PH} & LogRank
& 0.438 & 0.967 & 8.7 & 1.000
& 0.875 & 0.985 & 8.5 & 1.000
& 0.573 & 0.977 & 4.2 & 1.000 \\
& DetectGPT
& 0.160 & 0.741 & 8.7 & 1.000
& 0.000 & 0.459 & 8.5 & 1.000
& 0.000 & 0.543 & 4.2 & 1.000 \\
\arrayrulecolor{gray!40}\cline{3-14}\arrayrulecolor{black}
\multirow{2}{*}{LB} & KGW
& 0.995 & 0.999 & 11.9 & 0.628
& \textbf{1} & \textbf{1} & 11.0 & 0.682
& 0.620 & 0.990 & 4.8 & 0.792 \\
& SWEET
& \textbf{1} & \textbf{1} & 11.2 & \textbf{0.633}
& \textbf{1} & \textbf{1} & 10.2 & 0.685
& \textbf{0.890} & \textbf{0.995} & 4.6 & 0.791 \\
\arrayrulecolor{gray!40}\cline{3-14}\arrayrulecolor{black}
\multirow{2}{*}{DF} & DiPmark
& 0.975 & 0.999  &  \textbf{9.9} & 0.628
& 0.980 & 0.999 & \textbf{9.7} & \textbf{0.693}
& 0.550 & 0.964 & \textbf{4.4} & \textbf{0.797} \\
& EXP-edit
& 0.965 & 0.988 & 24.3 & 0.620
& 0.985 & \textbf{1} & 12.2 & 0.666
& 0.020 & 0.824 & 6.1 & 0.774 \\
\cline{2-14}
&\multicolumn{13}{>{\columncolor{blue!20}}c}{\scriptsize\textit{Summarization}} \\
\rowcolor{gray!20} \cellcolor{white} &  Un-watermarked
& --  & --  & 41.4 & 1.000
& -- & -- & 59.4 & 1.000
& -- & -- & 32.4 & 1.000 \\
\multirow{2}{*}{PH} & LogRank
& 0.050 & 0.303 & 41.4 & 1.000
& 0.000 & 0.195 & 59.4 & 1.000
& 0.281 & 0.767 & 32.4 & 1.000 \\
& DetectGPT
& 0.087 & 0.548 & 41.4 & 1.000
& 0.087 & 0.494 & 59.4 & 1.000
& 0.080 & 0.551 & 32.4 & 1.000 \\
\arrayrulecolor{gray!40}\cline{3-14}\arrayrulecolor{black}
\multirow{2}{*}{LB} & KGW
& 0.565 & 0.962 & 47.7 & 0.376
& 0.327 & 0.914 & 81.3 & 0.494
& \textbf{0.091} & \textbf{0.786} & 64.2 & 0.722 \\
& SWEET
& \textbf{0.820} & \textbf{0.985} & 45.2 & \textbf{0.410}
& 0.375 & 0.922 & 286.5 & 0.449
& 0.015 & 0.631 & 48.6 & \textbf{0.728} \\
\arrayrulecolor{gray!40}\cline{3-14}\arrayrulecolor{black}
\multirow{2}{*}{DF} & DiPmark
& 0.090 & 0.859  &  \textbf{41.7} & 0.393
& 0.083 & 0.741 & 115.9 & \textbf{0.504}
& 0.005 & 0.678 & 41.9 & 0.718 \\
& EXP-edit
& 0.015 & 0.949  & 65.8  & 0.393
& \textbf{0.960} & \textbf{0.994} & \textbf{20.1} & 0.402
& 0.005 & 0.528 & \textbf{33.2} & 0.724 \\
\hline
\end{tabular}
}
\caption{(RQ1) Detection and quality performance of six watermarking methods under three schemes: PH (\textit{post-hoc}), LB (\textit{logit-based}), and DF (\textit{distortion-free}) evaluated on three medical generation tasks with three different models. TPR was set at FPR = 0\%.} 
\label{tab:detection-quality}
\end{table*}

\subsection{Tasks and Datasets}
\label{subsec:tasks}

To assess watermark performance in realistic medical scenarios, we employ three distinct tasks using established medical datasets. First, \textbf{Text Completion} using the HealthQA dataset \cite{zhu2019hierarchical} requires models to complete medical passages (approx. 200 tokens), evaluating the impact on fundamental coherence and fluency. Second, \textbf{Question Answering (QA)}, also on HealthQA, challenges models to answer medical questions (approx. 200 tokens), specifically testing factual accuracy preservation under watermarking. Third, \textbf{Summarization} using the MeQSum dataset \cite{abacha2019summarization} involves generating concise summaries (< 100 tokens) of medical questions, evaluating performance on challenging short-text generation \cite{kirchenbauer2023watermark} and the crucial preservation of essential factual information. These tasks represent diverse medical tasks and output lengths, offering a comprehensive view of watermark impact on text quality and factual integrity in the medical domain. Details on how we built each task are available in Appendix \ref{sec:appendix_tasks}.

\subsection{Model and Hardware}
\label{subsec:model}
We use Meditron-7B \cite{chen2023meditron70b} as the main model for most experiments. This model included training on clinical guidelines, more suitable for consumer QA, unlike most models, which limited their training on abstracts and articles. Meditron is a medical adaptation model from Llama-2-7B \cite{touvron2023llama} through continued pretraining on the medical GAP-Replay corpus. We also evaluate other medical models like MedLlama-3-8B\footnote{https://hf.co/johnsnowlabs/JSL-MedLlama-3-8B-v2.0} and BioMistral 7B \cite{labrak2024biomistral}. Though this variation is not our main focus, we only include RQ1 results using these models and provide additional analysis in Appendix \ref{sec:appendix_model_analysis}. All experiments are conducted on a workstation equipped with an AMD Ryzen Threadripper PRO 3955WX (16-Cores) processor and two NVIDIA GeForce RTX 3090 GPUs (24GB VRAM each).

\section{Results}
We present our analysis of watermarking methods in answering a series of research questions defined in \S~\ref{subsec:rqs}. We begin by examining traditional automatic evaluation metrics in \S~\ref{subsec:rq1} (RQ1), comparing detectability measures against standard quality metrics to establish baseline performance. We then extend our investigation to task-specific metrics in \S~\ref{subsec:rq2} (RQ2), providing deeper insights regarding its resemblance of factuality aspects.

Building on the observed findings, we provide our Factuality-weighted Score (FWS) metrics in \S~\ref{subsec:rq3} (RQ3), which emphasize the factuality quality preservation. We validate the proposed metric through correlation analysis, comparing existing traditional automatic metrics and GPT-Judger assessments, and verify its alignment with human perception. Finally, we conduct an in-depth observation in \S~\ref{subsec:rq4}, to check the possible contributing factors to factuality corruption in watermarked texts (RQ4), offering insights for future watermarking method improvements.

\subsection{Detectability and Quality (RQ1)}
\label{subsec:rq1}
Table~\ref{tab:detection-quality} presents comparative performance of watermarking methods across generation tasks, measuring both detection capability and text quality preservation. \verb|TPR| measures detection rate, \verb|AUROC| overall robustness, \verb|PPL| text fluency, and \verb|SimCSE| semantic similarity. The results show a clear trade-off between detectability and output quality among different watermarking approaches.

Post-hoc methods (LogRank and DetectGPT) maintain the best possible text quality across all tasks, as they introduce no alterations to the generated text. However, this advantage comes at the cost of significantly reduced detectability. LogRank consistently outperforms DetectGPT across all tasks, reaching its highest detection performance in Question Answering (\verb|AUROC| 0.967), though its \verb|TPR| remains relatively low at 0.438. These low detectabilities are expected since no additional pattern is added to distinguish the generated text.

In contrast, generation-time methods demonstrate superior detection capabilities, with SWEET achieving perfect \verb|TPR| and \verb|AUROC| scores of 1 across Text Completion and Question Answering, and still best in Summarization. Quality-wise, DiPmark's distribution-preserving approach shows the best \verb|PPL| score across all tasks with a significant gap compared to other generation-time methods, though SWEET and KGW performed slightly better in \verb|SimCSE| score.

\begin{observation}
Post-hoc methods preserve text quality but suffer from weak detection, especially in \verb|TPR|. In contrast, \textbf{generation-time methods offer near-perfect detection with minimal quality loss}, making them a more balanced and promising approach.
\end{observation}
\vspace{1em}

To better understand how generation-time watermarking quality can be optimized from a downstream task perspective, the next subsection presents a focused evaluation of task-oriented performance.

\subsection{Task-oriented Performance (RQ2)}
\label{subsec:rq2}
Table~\ref{tab:rouge-f1} presents reference-based quality metrics evaluating generated text against natural text ground-truth, deliberately to give insights of factual preservation across different tasks. Unlike the general quality metrics in RQ1, task-specific metrics provide finer granularity \verb|ROUGE-2| measures bigram overlap between generated and reference text, \verb|ROUGE-L| captures the longest common subsequence to reflect order matching, \verb|F1| balances precision and recall to quantify overall matching accuracy and \verb|AlignScore| evaluates factual consistency. Higher values across all metrics indicate better performance.

\begin{table}[h]
\centering
\footnotesize
\resizebox{\columnwidth}{!}{%
\begin{tabular}{lccccccccc}
\hline
\multirow{2}{*}{\textbf{\makecell[l]{Waterm. \\ Methods}}} & 
\multicolumn{3}{c}{\textbf{\makecell[c]{Text \\ Completion}}} & 
\multicolumn{3}{c}{\textbf{\makecell[c]{Question \\ Answering}}} & 
\multicolumn{3}{c}{\textbf{Summarization}} \\
\cline{2-10}
 & \tiny RG-2 & \tiny RG-L & \tiny AS & \tiny RG-2 & \tiny F1 & \tiny AS  & \tiny RG-2 & \tiny RG-L & \tiny AS \\
\hline
\rowcolor{gray!20}
Un-waterm. & .038 & .142  & .241  & .027 & .130 & .264 & .046 & .156 & .114 \\
KGW        & .030 & .135 &\textbf{ .255} & .021 & \textbf{.127} & \textbf{.273} & .025 & .125 & .096 \\
SWEET      & .029  & .132 & .227 & .022 & .125 & .235 & .032 & .128 & \textbf{.121} \\
DiPmark    & \textbf{.033} & \textbf{.137} & .216 & .021 & .125 & .238 & .046 & \textbf{.153} & .110 \\
EXP-edit   & .029 & .129 & .223 & \textbf{.023}  & .123 & .254 & \textbf{.049} & .142 & .104 \\
\hline
\end{tabular}%
}
\caption{(RQ2) Watermarking methods Task-oriented performance across various tasks. RG* stands for ROUGE* and AS stands for AlignScore.}
\label{tab:rouge-f1}
\end{table}

\begin{figure}[t]
\centering
  \includegraphics[width=0.8\linewidth]{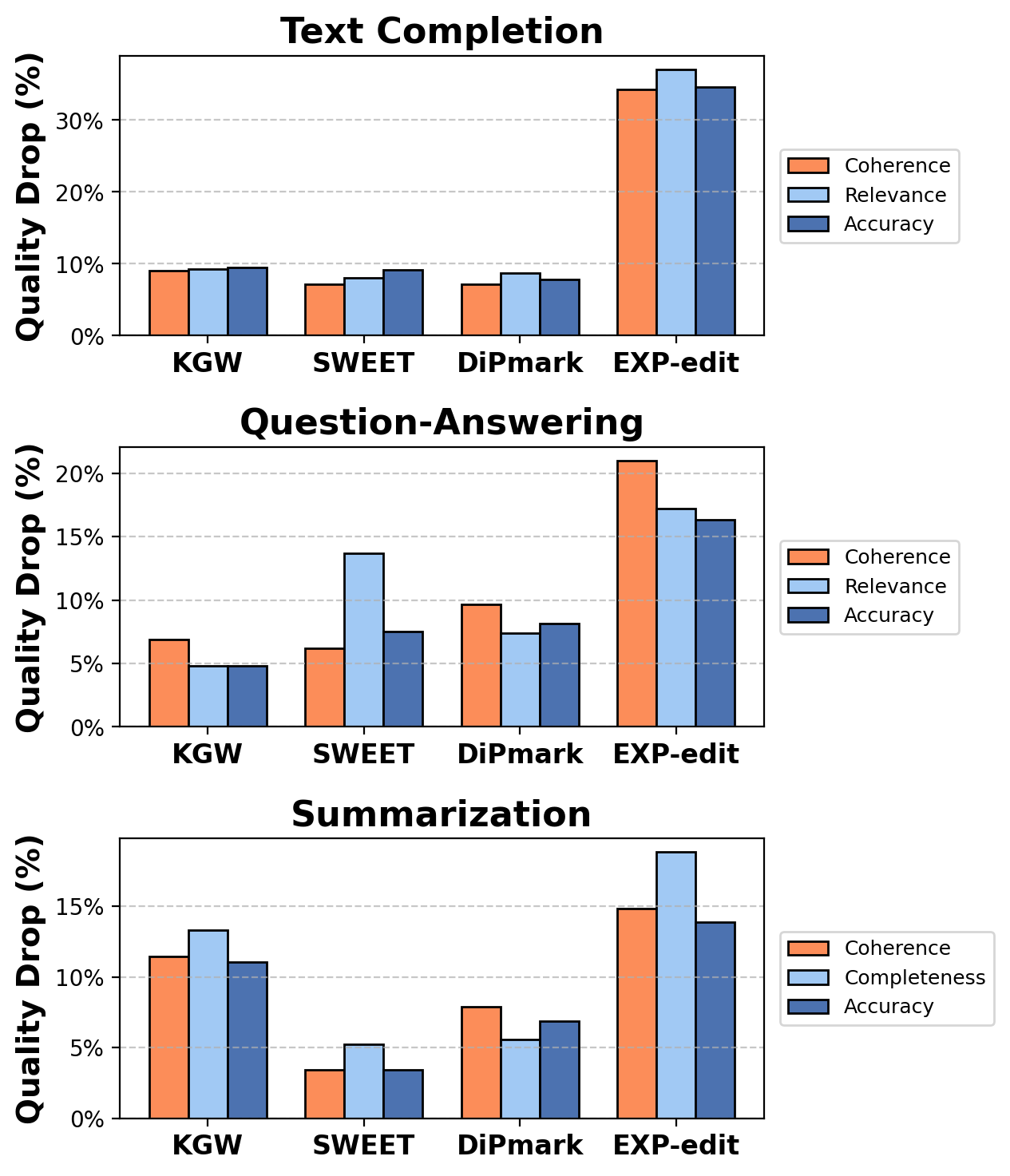} 
  \caption {(RQ3) GPT-Judger quality aspects cover the dimensions of \emph{coherence}, \emph{relevance}, and \emph{factual accuracy}, showing degradation across watermarking methods. On average, relevance and factual accuracy suffer greater degradation than coherence.}
  \label{fig:gptjudge-result}
\end{figure}

In Text Completion, DiPmark achieved the best performance with \verb|ROUGE-2| (0.033) and \verb|ROUGE-L| (0.137), slightly above other watermarking methods, while getting the lowest \verb|AlignScore| (0.216). For Question Answering, KGW gets the best performance with the highest \verb|F1| score (0.127) and \verb|AlignScore| (0.273), though all methods show minimal degradation from un-watermarked text. In Summarization, EXP-edit achieves \verb|ROUGE-2| (0.049) that exceeds even the un-watermarked text (0.046), likely because its alternative token sampling approach modifies probability distributions rather than directly applying specific token selections.

\begin{observation}
    DiPmark demonstrates strong performance in Text Completion, while EXP-edit notably exceeds un-watermarked quality in Summarization. However, overall differences between methods are marginal (<1\%), indicating \textbf{task-specific metrics are insufficient for drawing clear factuality quality conclusions}. 
\end{observation}
Addressing traditional automatic metrics' limitation, the next subsection (RQ3) introduces GPT-Judger and our Factuality-Weighted Score (FWS) for a more tailored, factuality-focused evaluation in the medical domain.

\subsection{Factuality-weighted Score FWS (RQ3)}
\label{subsec:rq3}
\textbf{GPT-Judger quality aspects}\hspace{0.5em}
Figure ~\ref{fig:gptjudge-result} shows results of GPT-Judger Quality aspects defined in Table ~\ref{tab:quality-aspects}, evaluating coherence (orange) and factuality dimensions (blue). The results highlight quality degradations when watermarking techniques are applied to language models, ranging between 3\%-37\% across tasks. This pattern is most pronounced with EXPEdit in Text Completion, showing significant degradations (34-37\% drops) across all dimensions. Another interesting sight is in the Question Answering task, where SWEET's relevance quality degrades 2x the coherence's degradation.

Additionally, by averaging percentage drops across all tasks and methods, we observe that relevance suffers the most substantial degradation at approximately 12.42\%, compared to coherence's average drop of 11.58\% and accuracy's 11.09\%. Though the difference seems not significant, Figure \ref{fig:gptjudge-result} shows contradictory relations (SWEET on QA, DiPmark on Summarization), where eliminating one over the other aspect will not be insightful. It is also worth considering that the drop in accuracy aspect itself indicates factuality corruption, as the Judger agrees that it introduces inaccurate medical terms, potentially signaling hallucinations. These findings indicate that coherence alone is inadequate and can be misleading. While a model's output may maintain reasonable coherence, its factual reliability may experience more compromises. Based on these insights, we incorporate these detailed assessments into a unified Factuality-weighted Score (FWS).\\

\noindent\textbf{FWS of Automatic metrics and GPT-Judger}\hspace{0.5em} The results of unified quality metrics using FWS can be seen in Table ~\ref{tab:factuality-weighted} through both traditional automatic metrics \verb|(Auto)| and GPT-Judger \verb|(GPTJ)| quality assessment. FWS of GPT-Judger metrics provides more interpretable evaluation of watermarking methods, with wider score differentials (ranging from 0.369 to 0.556) compared to FWS using traditional automatic metrics, which are tighter (0.135 to 0.195). This distinction provides a clearer assessment of the performance. 

\begin{table}[h]
\centering
\footnotesize
\resizebox{\columnwidth}{!}{%
\begin{tabular}{lcccccc}
\hline
\multirow{2}{*}{\textbf{\makecell[l]{Watermarking \\ Methods}}} & 
\multicolumn{2}{c}{\textbf{\makecell[c]{Text \\ Completion}}} & 
\multicolumn{2}{c}{\textbf{\makecell[c]{Question \\ Answering}}} & 
\multicolumn{2}{c}{\textbf{Summarization}} \\
\cline{2-7}
 & \tiny Auto & \tiny GPTJ & \tiny Auto & \tiny GPTJ & \tiny Auto & \tiny GPTJ \\
\hline
KGW & 0.195 & 0.540 & {0.185} & {0.411} & 0.135 & 0.419 \\
SWEET & 0.193 & {0.556} & {0.185} & 0.393 & 0.146 & {0.451} \\
DiPmark & {0.197} & 0.552 & 0.184 & 0.408 & {0.158} & 0.432 \\
EXP-edit & 0.188 & 0.369 & 0.183 & 0.369 & {0.155} & 0.405 \\
\hline
\end{tabular}%
}
\caption{(RQ3) Factuality-weighted score (FWS) across watermarking methods, generation tasks and evaluation schemes. FWS using GPT-Judger aspects \texttt{(GPTJ)} presents clearer distinctions than traditional automatic metrics \texttt{(Auto)}. }
\label{tab:factuality-weighted}
\end{table}

For the task-specific performance across watermarking approaches: SWEET demonstrates superior performance in Text Completion (0.556), while KGW achieves the highest scores in both Question Answering (0.411) and Summarization (0.419). EXP-edit consistently shows the weakest performance across all tasks. This is likely due to its alternative token sampling strategy, which causes the model to behave almost like a different language model compared to the un-watermarked version. In this case, it introduces more drawbacks than the intended benefit of distortion-free watermark, as claimed \cite{kuditipudi2024robust}.

Another interesting result in the Question Answering task is that SWEET and KGW scores are the same under automatic metrics, but SWEET falls behind in the GPT-Judger result. This is due to greater factuality degradation, as supported in Figure~\ref{fig:gptjudge-result}. Since factuality is weighted more in FWS, SWEET's score behind KGW's by about 2\%. Overall, indicating the reliability of our proposed FWS. Next, we calculate the correlation of both FWS (\verb|Auto| and \verb|GPTJ|) and human ratings to assess alignment with human judgment.

\begin{table}[h]
\small
  \centering
  \resizebox{\columnwidth}{!}{
  \begin{tabular}{lccc}
    \hline
    \textbf{Configuration} & \makecell[c]{\textbf{FWS Param.} \\ \textbf{($\alpha$, $\beta$)}} & \makecell[c]{\textbf{Auto-} \\ \textbf{Human}} & \makecell[c]{\textbf{GPTJ-} \\ \textbf{Human}} \\
    \hline
    \makecell[l]{Coherence-Heavy  (1:2)}   & \makecell[c]{(0.25, 0.5)}   & 0.089 & 0.814 \\
    \makecell[l]{Equal Weighting  (1:1)}   & \makecell[c]{(0.33, 0.33)}  & 0.181 & 0.833 \\
    \makecell[l]{Current (Ours)  (2:1)}    & \makecell[c]{(0.4, 0.2)}    & \textbf{0.248} & \textbf{0.840} \\
    \makecell[l]{Factuality-Heavy  (4:1)}  & \makecell[c]{(0.44, 0.11)}  & 0.276 & 0.841 \\
    \makecell[l]{Factuality-Heavy  (6:1)}  & \makecell[c]{(0.46, 0.08)}  & 0.283 & 0.841 \\
    \hline
  \end{tabular}}
  \caption{Sensitivity analysis comparing correlations of automatic and GPT-Judger with human evaluation under different weighting configurations, with stronger correlations observed when factuality is emphasized.}
  \label{tab:config-correlation}
\end{table}

\noindent\textbf{Human Evaluation}\hspace{0.5em}
Human ratings were evaluated using the Nemenyi test and p-value significance on quality aspects defined in Table~\ref{tab:quality-aspects}. The p-values for \verb|(Coherence, Relevance)|, \verb|(Coherence, FactualAccuracy)|, and \verb|(Relevance, FactualAccuracy)| are 0.635, 0.001, and 0.004, indicating clear distinctions among the aspects. \textbf{These quality aspects exhibit distinct characteristics} consistent with intuition, highlighting the importance of detailed factuality aspects in the medical domain. Sensitivity analysis of Equation \ref{eq:factuality-weighted} shows \textbf{a consistent trend: greater emphasis on factuality improves correlation with human judgment}. While the 2:1 weighting did not yield the highest correlation, it outperformed coherence-heavy settings and was chosen as an intuitive benchmark for a factuality-focused configuration (Table \ref{tab:config-correlation}).

\begin{table}[h]
\small
  \centering
  \begin{tabular}{lcccc}
    \hline
   \textbf{Metrics} & \textbf{Coh.} & \textbf{Rel.} & \textbf{FactAcc.} & \textbf{FWS} \\
    \hline
    Auto &  0.070& 0.355 & 0.497 & 0.256  \\
    GPTJ &  \textbf{0.701}&  \textbf{0.881} &  \textbf{0.613} & \textbf{0.839} \\
    \hline
  \end{tabular}
  \caption{(RQ3) Pearson correlation of Human evaluation in QA task. GPT-Judger (\texttt{GPTJ}) correlated strongly for each aspect and aligned better to human evaluation than traditional automatic metrics (\texttt{Auto}).}
  \label{tab:human-correlation}
\end{table}

Consequently, Table ~\ref{tab:human-correlation} shows the Pearson correlation of human evaluation in the Question Answering task. The correlation for each aspect ranging from 0.6 to 0.8, indicate that  \textbf{human ratings aligned closer to GPT-Judger (\texttt{GPTJ})} than traditional automatic metrics (\texttt{Auto}), provided higher absolute values indicate stronger correlations. 

Taken together, these analyses offer some key observations regarding the impact of watermarking on factuality, FWS reliability, and human alignment.
\begin{observation}
\begin{itemize}[leftmargin=*]
    \item Watermarking techniques introduce \textbf{significant factuality degradation}, with EXPEdit showing the most severe drops (up to 37\%).
    \item \textbf{Factuality-weighted Scores (FWS)} based on GPT-Judger \textbf{offer clearer and more interpretable distinctions} than traditional automatic metrics.
    \item \textbf{Human evaluations confirm the distinct role of factuality dimensions beyond coherence} and show strong alignment with GPT-Judger, \textbf{validating the proposed evaluation framework}.
\end{itemize}
\end{observation}
\vspace{1em}

\subsection{Risk of Factuality Corruption (RQ4)}
\label{subsec:rq4}
The previous subsection highlighted the degradation of factuality quality dimensions caused by watermarking techniques. In this subsection, we further investigate the potential contributing factors behind this quality drop. To ensure a fair analysis, we use natural text from the datasets as the reference, assuming it represents factuality from medical domain expertise.\\

\begin{figure}[tbp]
\includegraphics[width=\columnwidth]{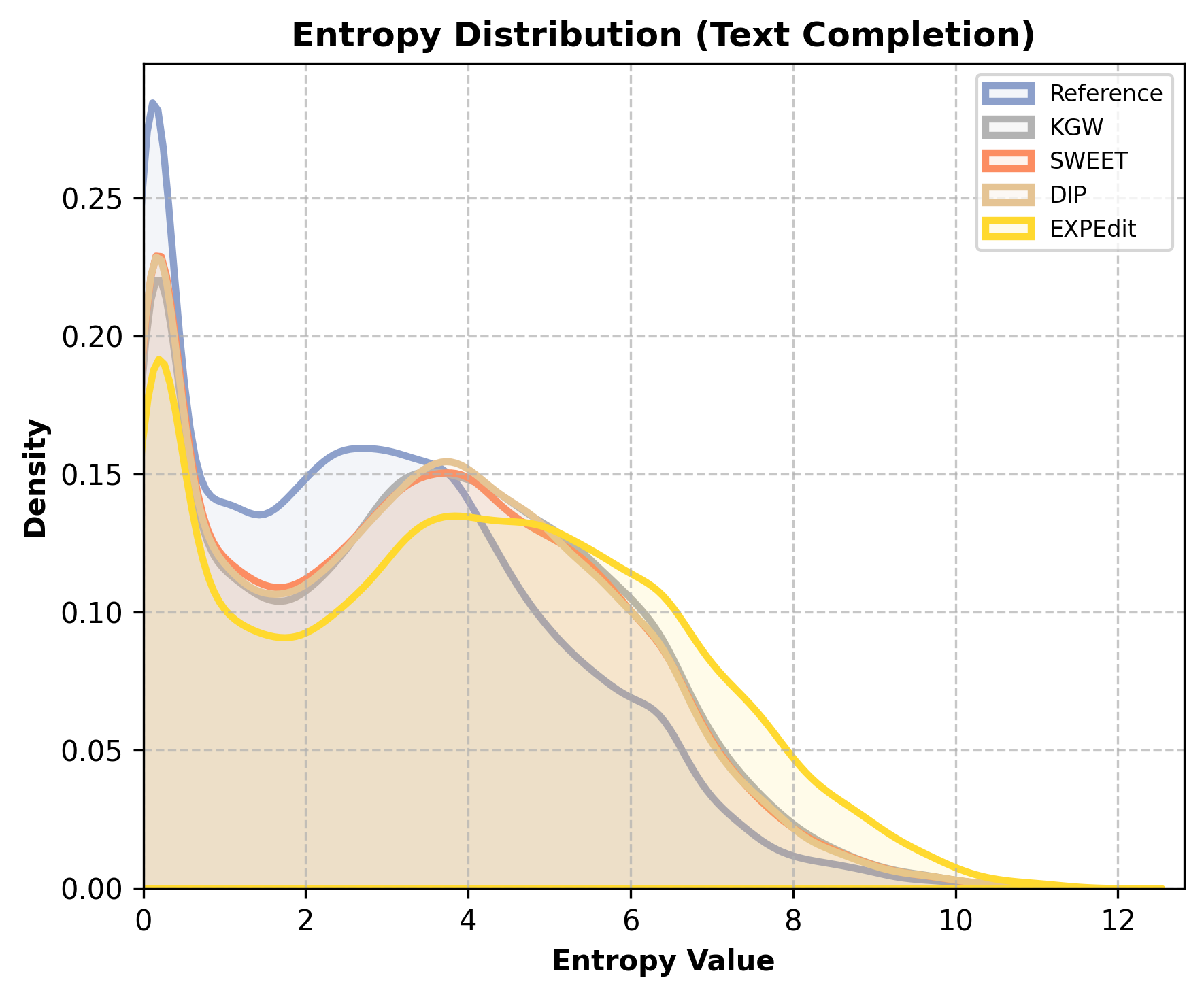}
\caption{(RQ4) All watermarking methods cause token entropy distribution shifts, especially on the low-entropy side where two hills appear. EXPEdit shows the biggest change overall.}
\label{fig:plot-entropy}
\end{figure}

\noindent\textbf{Token entropy distribution}\hspace{0.5em} In Figure~\ref{fig:plot-entropy}, we observe a distinct pattern: the reference text’s entropy is primarily concentrated in low-entropy regions. While the watermarked texts follow a similar overall shape, their \textbf{density shifts by approximately 10\%, spreading across both low and mid-entropy} areas. Similar pattern of low-entropy distribution shifts for Question Answering and Summarization described in Appendix \ref{sec:appendix_risk}. Given that this shift may have a non-trivial impact, specifically in the second entropy range ($\text{1.5--4}$), which hypothetically corresponds to medical terminology, we conduct a deeper analysis of low-entropy tokens and their types to assess the potential risk of reduced model confidence in predicting critical terms.

\begin{figure}[htbp]
\includegraphics[width=\columnwidth]{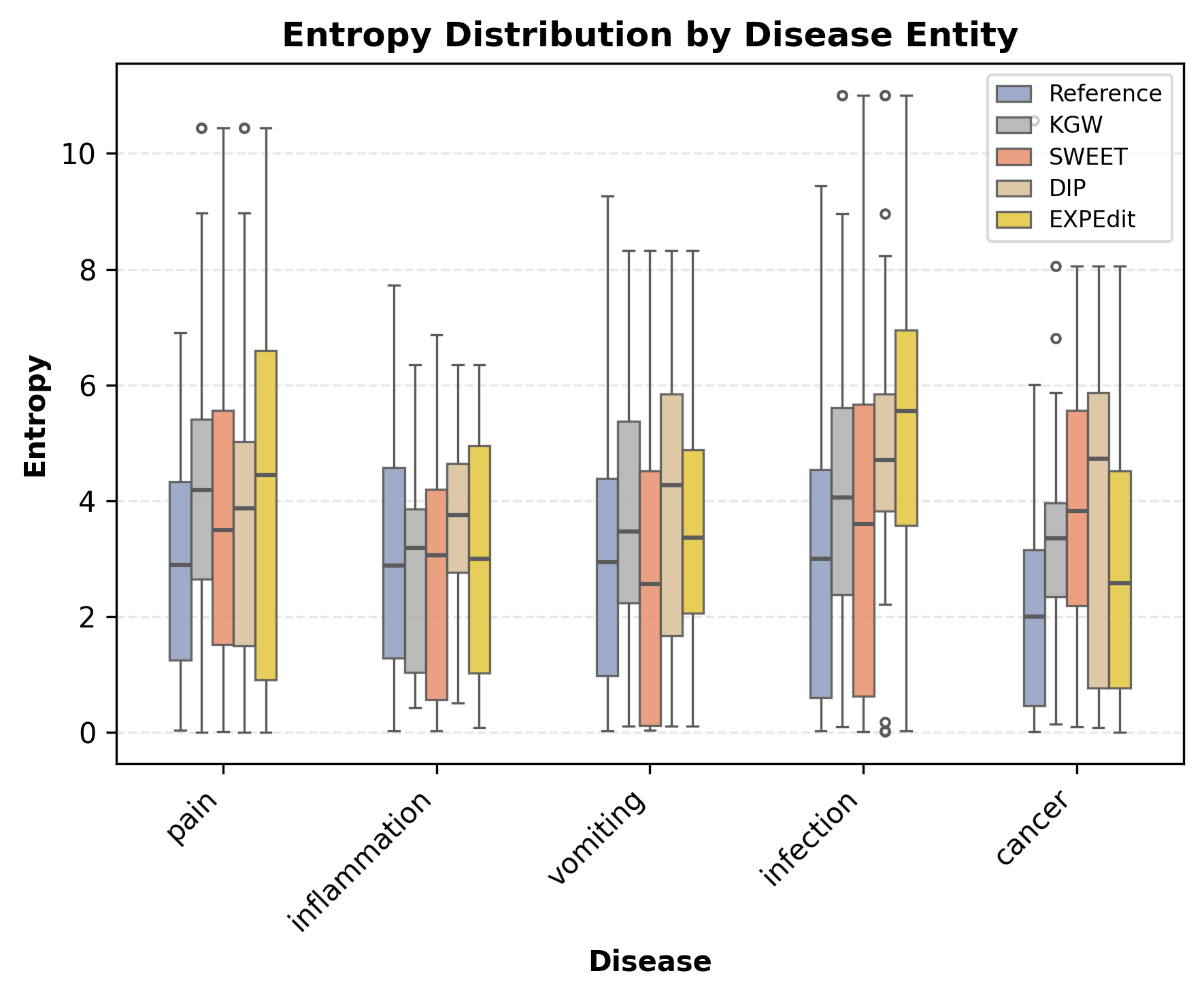}
\caption{(RQ4) Entropy distribution by most frequent disease entities. Median shifts toward higher entropy reflect reduced contextual confidence in generating medical terminology.}
\label{fig:plot-boxplot_entity}
\end{figure}

\noindent\textbf{Entropy distribution by disease entity}\hspace{0.5em} As shown in Figure~\ref{fig:plot-boxplot_entity}, the reference token entropy for disease entities \cite{neumann-etal-2019-scispacy} generally aligns with the second entropy range ($\text{2--4}$) in Figure ~\ref{fig:plot-entropy}. Within this range, we observe noticeable shifts in median token entropy across disease entities shown in the plot, such as \emph{pain}, \emph{infection}, and \emph{cancer}. These shifts suggest a \textbf{degradation in the model’s contextual confidence when handling critical domain-specific terms}. It also indicates that the more uncertain a model is on a disease entity, the more likely it is to hallucinate about that entity. This intuition will further be observed in a more detailed entity-level comparison. \\

\noindent\textbf{Entity hallucinations}\hspace{0.5em} 
Inspired by LLMs factuality hallucination described by \citet{li-etal-2024-dawn}, which categorizes erroneous entities as entity-error hallucination, we analyze factuality corruption in watermarked medical texts. Using a disease entity recognizer \cite{neumann-etal-2019-scispacy}, we measure \textit{Introduced entities} (new disease entities not found in reference text) and \textit{Hallucination rate} (percentage of semantically divergent entities). We compute pairwise contextual BERT embeddings with a cosine similarity \cite{zhangbertscore} threshold of 0.6, which allows us to distinguish acceptable synonyms from true semantic shifts that corrupt the text's medical context.

\begin{table}[h]
\small
  \centering
  \begin{tabular}{lcc}
    \hline
   \textbf{\makecell[l]{Watermarking \\ Methods}} & \textbf{\makecell[l]{Avg.\\ Introduced \\ Entities}} & \textbf{\makecell[l]{Increased \\ Hallucination \\ rate (\%)}}  \\ 
   \hline
    KGW &  4.13 &  3.1 \\
    SWEET &  4.39 &  3.2 \\
    DiPmark &  4.10 & \textbf{0.6} \\
    EXP-edit &  \textbf{3.63} &  1.8 \\
    \hline
  \end{tabular}
  \caption{(RQ4) Semantic corruption in watermarked texts measured by new entity introduction and hallucination rates. New disease entities occupy up to 11-13\% of content and produce varying degrees of semantic distortion (0.6-3.2\%), revealing how watermarking risks factual integrity in the medical domain.}
  \label{tab:entity-hallucination}
\end{table}

Table~\ref{tab:entity-hallucination} presents entity hallucination statistics across watermarking methods. On average, each watermarking technique introduces 3.6-4.4 new disease entities per 200 tokens of text, with each entity typically spanning about 6 tokens, meaning up to 11-13\% of watermarked content consists of introduced medical terms. The increased hallucination rate, on the other hand, shows that \textbf{all methods led to higher hallucination rates compared to un-watermarked texts, ranging from 0.6\% to 3.2\%}. Samples of the hallucinated text can be found in Appendix \ref{sec:halu}. These findings suggest watermarking methods risk their semantic preservation capabilities by introducing entity hallucinations, which further support the quality degradation result in the previous result Figure~\ref{fig:gptjudge-result}.\\

\begin{observation}
\begin{itemize}[leftmargin=*]
    \item \textbf{Entropy shift}: token entropy distribution shifts by ~10\% in low and mid-entropy regions, disrupting natural language patterns.
    \item \textbf{Entity confidence}: reduced model confidence when handling disease entities, potentially correlated with hallucination.
    \item \textbf{Entity hallucination}: new disease entities introduced covering 11-13\% of content and increased hallucination rates by up to 3.2\%, creating semantic corruption of medical factual information.
\end{itemize}
\end{observation}

\section{Conclusion}
Our evaluation of LLM watermarking methods for medical texts reveals a critical trade-off: while current approaches achieve high detection capabilities, they compromise factual integrity. The proposed Factuality-Weighted Score addresses this concern by prioritizing factual accuracy over coherence, validated through human evaluations. Our analysis shows watermarking shifts token distributions, affecting medical terminology and introducing entity hallucinations.

These findings highlight the need for domain-aware watermarking techniques that preserve medical content integrity while maintaining detectability. We recommend developing approaches that specifically account for medical terminology alongside more robust factuality-focused evaluation protocols. As LLMs continue to be deployed in healthcare settings, ensuring content authenticity and factual reliability remains important.

\section*{Limitations}
This study includes a human evaluation to validate the alignment between GPT-Judger assessments and human judgments, restricted to the Question Answering (QA) task. Our objective was not to establish comprehensive human evaluation across all tasks, but to assess the reliability of our proposed workflow, which we demonstrate through a strong correlation with human judgments in the QA setting. We also identified contributing factors to factuality degradation, such as entropy distribution shifts and entity hallucinations, while other potential factors likely exist beyond our analysis. The connections between these mechanisms and various watermarking approaches represent a promising area for further research toward developing more domain-sensitive watermarking techniques.

\section*{Ethical Statement}
This work investigates the implementation of existing LLM watermarking methods in medical texts. We note that applying these methods without consideration of domain-specific factuality may contribute to misinformation. While this study provides high level insights of factuality concerns, further refinement to different use case might be necessary to ensure safe and responsible deployment in real-world medical applications.

\bibliography{custom}
\newpage
\appendix
\section{Additional Analysis}
\label{sec:appendix_ANALYSIS}

\subsection{Model Analysis}
\label{sec:appendix_model_analysis}

Table~\ref{tab:model-scaling} presents a comparison of watermarking methods across medical language models for text QA tasks: Meditron 7B\footnote{https://hf.co/epfl-llm/meditron-7b}, MedLlama-3 8B\footnote{https://hf.co/johnsnowlabs/JSL-MedLlama-3-8B-v2.0} and BioMistral 7B\footnote{https://hf.co/BioMistral/BioMistral-7B}. High detection metrics (TPR=1, AUROC=1) can be seen on smaller models (Meditron 7B and BioMistral 7B), with a decrease observed in the larger MedLlama-3 8B model. As expected, quality metrics improve with model size, with MedLlama-3 8B showing substantially higher SimCSE scores compared to smaller models, and better task performance in all metrics.

\begin{table}[h]
\centering
\footnotesize
\resizebox{0.7\columnwidth}{!}{
\begin{tabular}{lccc}
\hline
\multirow{2}{*}{\small \textbf{\makecell[l]{Watermarking \\Methods}}} &  \multicolumn{3}{c}{\small \textbf{Task}}\\
\cline{2-4}
 & \scriptsize \textbf{RG-2}$\uparrow$  & \scriptsize \textbf{F1}$\uparrow$ & \scriptsize \textbf{AS}$\uparrow$  \\
\hline
&\multicolumn{3}{>{\columncolor{blue!20}}c}{Meditron v1.0 7B} \\
KGW & .021 & \textbf{.127} & \textbf{.273 }\\
SWEET & .022 & .125 & .235 \\
DiPmark & .021 & .125 & .254\\
EXP-edit & \textbf{.023} & .123 & .238 \\
&\multicolumn{3}{>{\columncolor{blue!20}}c}{BioMistral 7B} \\
KGW & .026 & .135 & \textbf{.311} \\
SWEET & .028 & .137 & .286\\
DiPmark & \textbf{.030} & \textbf{.141}  & .297\\
EXP-edit & .029 & .135 & .297 \\
&\multicolumn{3}{>{\columncolor{blue!20}}c}{MedLlama-3 8B} \\
KGW & .042 & \textbf{.154} & .364 \\
SWEET & .045 & \textbf{.154} & \textbf{.387}\\
DiPmark  & .045 & \textbf{.154} & .373 \\
EXP-edit & \textbf{.048} & .153 & .375\\
\hline
\end{tabular}
} 
\caption{Performance comparison of watermarking methods across medical language models of different sizes for QA tasks.}
\label{tab:model-scaling}
\end{table}

When evaluating the quality using GPT-Judger, we still can see that the more recent models, BioMistral 7B and MedLlama-3 8B, still experience a quality drop as shown in Table~\ref{tab:model-scaling-2}. 

\begin{table}[h]
\centering
\footnotesize
\resizebox{0.9\columnwidth}{!}{
\begin{tabular}{lrrr}
\hline
\multirow{2}{*}{\small \textbf{\makecell[l]{Watermarking \\Methods}}}  &  \multicolumn{3}{c}{\small \textbf{GPT-Judger Quality Drop (\%)}}\\
\cline{2-4}
 & \scriptsize \textbf{Coherence}$\downarrow$  & \scriptsize \textbf{Relevance}$\downarrow$ & \scriptsize \textbf{Factual Accuracy}$\downarrow$  \\
\hline
&\multicolumn{3}{>{\columncolor{blue!20}}c}{Meditron v1.0 7B} \\
KGW & 6.9 & \textbf{4.8}  & \textbf{4.8} \\
SWEET & \textbf{6.2} & 13.7  & 7.5 \\
DiPmark & 9.6 & 7.4  & 8.1 \\
EXP-edit & 20.9 & 17.2  & 16.3 \\
&\multicolumn{3}{>{\columncolor{blue!20}}c}{BioMistral 7B} \\
KGW & 12.4 & 12.8  & 12.1 \\
SWEET & 8.5 & 9.2  & 6.6 \\
DiPmark & \textbf{3.3} & \textbf{5.9}  & \textbf{6.0} \\
EXP-edit & 17.9 & 24.1   & 15.9 \\
&\multicolumn{3}{>{\columncolor{blue!20}}c}{MedLlama-3 8B} \\
KGW & 7.5 & 9.7  & \textbf{3.7} \\
SWEET & 9.5 & 15.0  & 11.9 \\
DiPmark & \textbf{6.9} & \textbf{5.3}  & 5.3  \\
EXP-edit & 7.5 & 5.8   & 6.3 \\
\hline
\end{tabular}
} 
\caption{Performance comparison of watermarking methods across medical language models of different sizes for QA tasks.}
\label{tab:model-scaling-2}
\end{table}

\subsection{Hyperparameter Effect on Detection}
\label{sec:appendix_hyperparameter}

The parameters used for each watermarking method are shown in Table \ref{tab:defaultparam}. These are the default parameters used in MarkLLM\footnote{https://github.com/THU-BPM/MarkLLM} based on each original paper. For details on the usage of each parameter, please refer to the original papers.

\begin{table}[h]
\small
  \centering
  \begin{tabular}{ll}
    \hline
    \textbf{\makecell[ll]{Watermarking \\ Methods}} & \textbf{Parameters}  \\
    \hline
    KGW & \makecell[ll]{$\gamma=0.5, \delta=2.0$}\\
    SWEET & \makecell[ll]{$\gamma=0.5, \delta=2.0$, \\ $\text{entropy\_threshold}=0.9$} \\
    DiPmark & \makecell[ll]{$\gamma=0.5, \alpha=0.45$}\\
    EXP-edit & \makecell[ll]{$\text{pseudo\_length}=256$}\\
    \hline
  \end{tabular}
  \caption{Hyperparameters used in main experiments for each watermarking method.}
  \label{tab:defaultparam}
\end{table}

\begin{figure}[h]
    \centering
    \begin{subfigure}{0.45\textwidth}
        \centering
        \includegraphics[width=\textwidth]{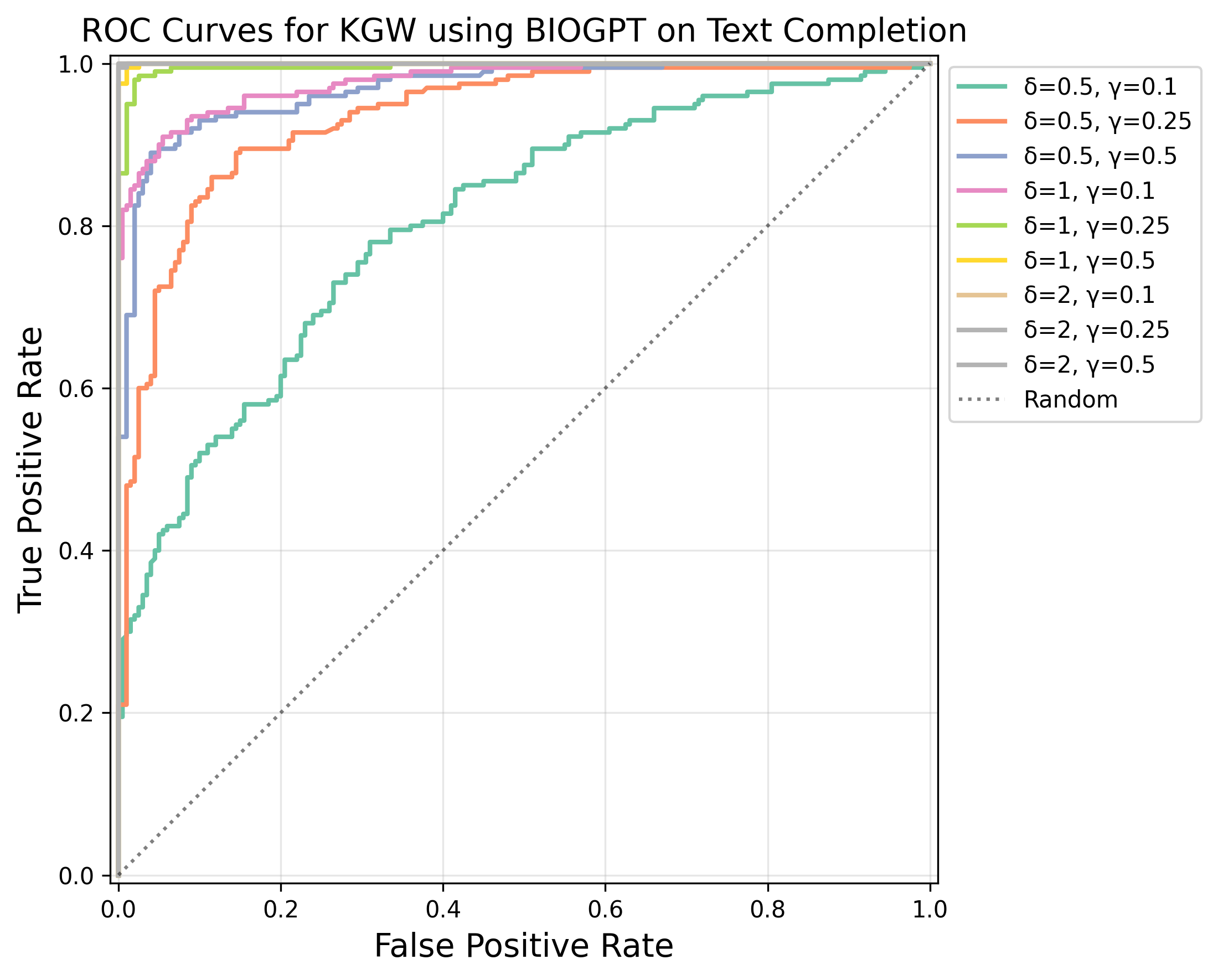}
        \label{fig:roc-sweet}
    \end{subfigure}
    \hfill
    \begin{subfigure}{0.45\textwidth}
        \centering
        \includegraphics[width=\textwidth]{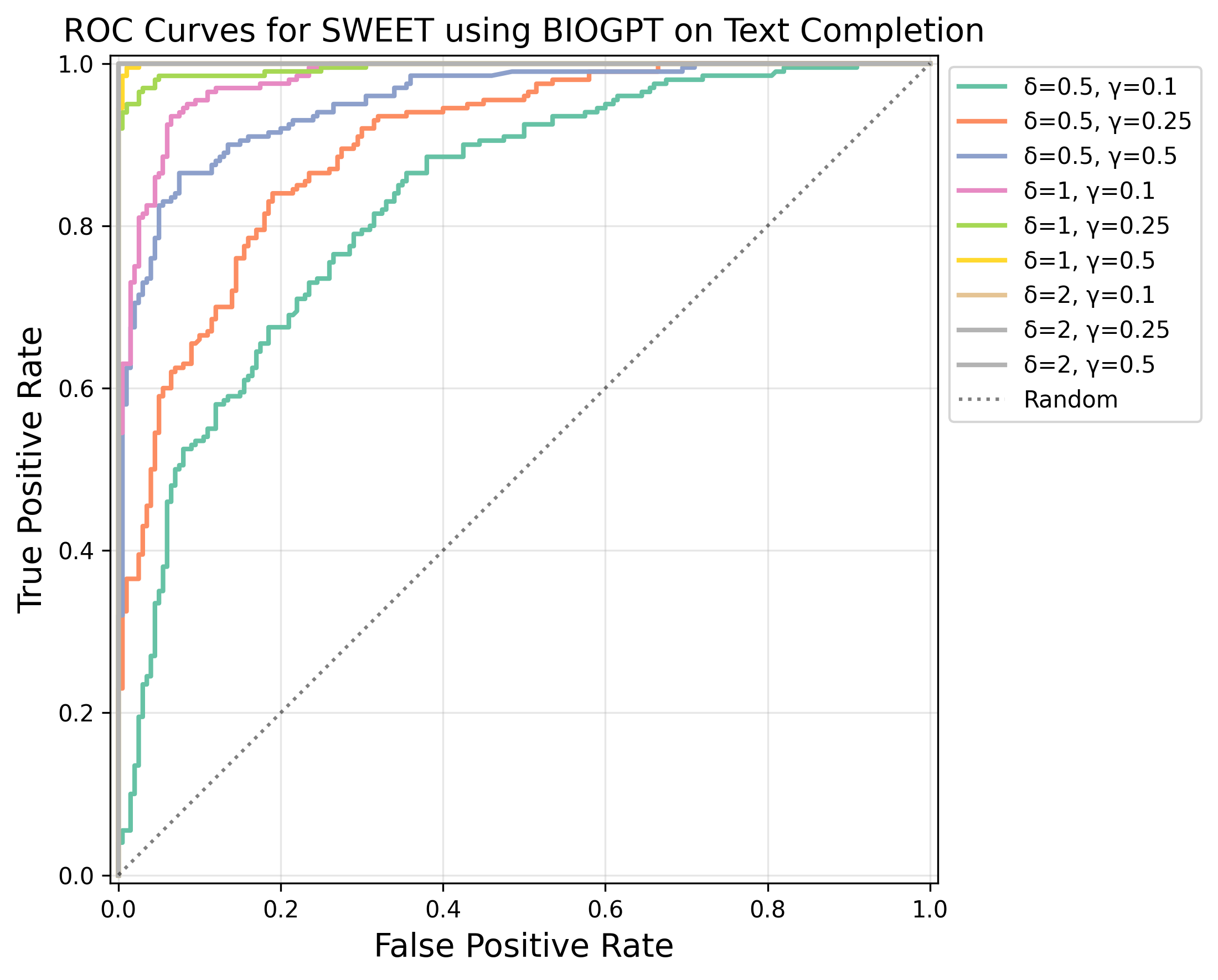}
        \label{fig:roc-kgw}
    \end{subfigure}
    \caption{ROC Curve for KGW (top) and SWEET (bottom) with different $\delta$ and $\gamma$ in Text Completion task.}
    \label{fig:roc-plots}
\end{figure}

Additionally, we conducted a systematic investigation into how hyperparameter values affect detection performance. First, we explored the impact of varying $\delta \in \{0.5, 1, 2\}$ and $\gamma \in \{0.1, 0.25, 0.5\}$ in logit-based watermarking, KGW, and SWEET, with results presented in Figure~\ref{fig:roc-plots}. For these experiments, we utilized BioGPT\footnote{https://hf.co/microsoft/biogpt} \cite{luo2022biogpt} model, which is computationally less demanding than the larger models used in our main analysis, enabling more efficient hyperparameter exploration. Our findings demonstrate that for both KGW and SWEET schemes, larger values of $\delta$ and $\gamma$ correspond to improved detection rates. However, as discussed in \S~\ref{subsec:rq1}, there exists a fundamental trade-off between detection efficacy and the quality of generated content.

\begin{table}[h]
\small
  \centering
  \begin{tabular}{lcc}
    \hline
    EXP-edit parameters & \textbf{TPR} & \textbf{AUROC} \\
    \hline
    $\text{pseudo\_length}=100$ & \textbf{0.890} & \textbf{0.974} \\
    $\text{pseudo\_length}=200$ & 0.875 & 0.938 \\
    $\text{pseudo\_length}=300$ & 0.830 & 0.949 \\
    \hline
  \end{tabular}
  \caption{TPR and AUROC for EXP-edit with different pseudo length ($n$) in Text Completion task}
  \label{tab:hiperparamexpedit}
\end{table}

\begin{table}[h]
\small
  \centering
  \begin{tabular}{lcc}
    \hline
    DiPmark parameters & \textbf{TPR} & \textbf{AUROC} \\
    \hline
    $\alpha=0.4$ & 0.985 & 0.990 \\
    $\alpha=0.45$ &\textbf{ 0.995} & \textbf{0.999} \\
    $\alpha=0.5$ & 0.980 & \textbf{0.999} \\
    \hline
  \end{tabular}
  \caption{TPR and AUROC for DiPmark with different $\alpha$ threshold in Text Completion task}
  \label{tab:hiperparamdipmark}
\end{table}

Hyperparameter exploration for EXP-edit and DiPmark is shown in Table \ref{tab:hiperparamexpedit} and Table \ref{tab:hiperparamdipmark}, respectively. We can see that EXP-edit achieves optimal performance with shorter pseudo lengths ($n=100$), yielding \verb|TPR| of 0.890 and \verb|AUROC| of 0.974. While DiPmark demonstrates consistently higher detection rates with its best configuration at $\alpha=0.45$, achieving near-perfect results (\verb|TPR|=0.995, \verb|AUROC|=0.999).

\subsection{Risk of Factuality Corruption}
\label{sec:appendix_risk}
\begin{figure}[h]
    \centering
    \begin{subfigure}{0.42\textwidth}
        \centering
        \includegraphics[width=\textwidth]{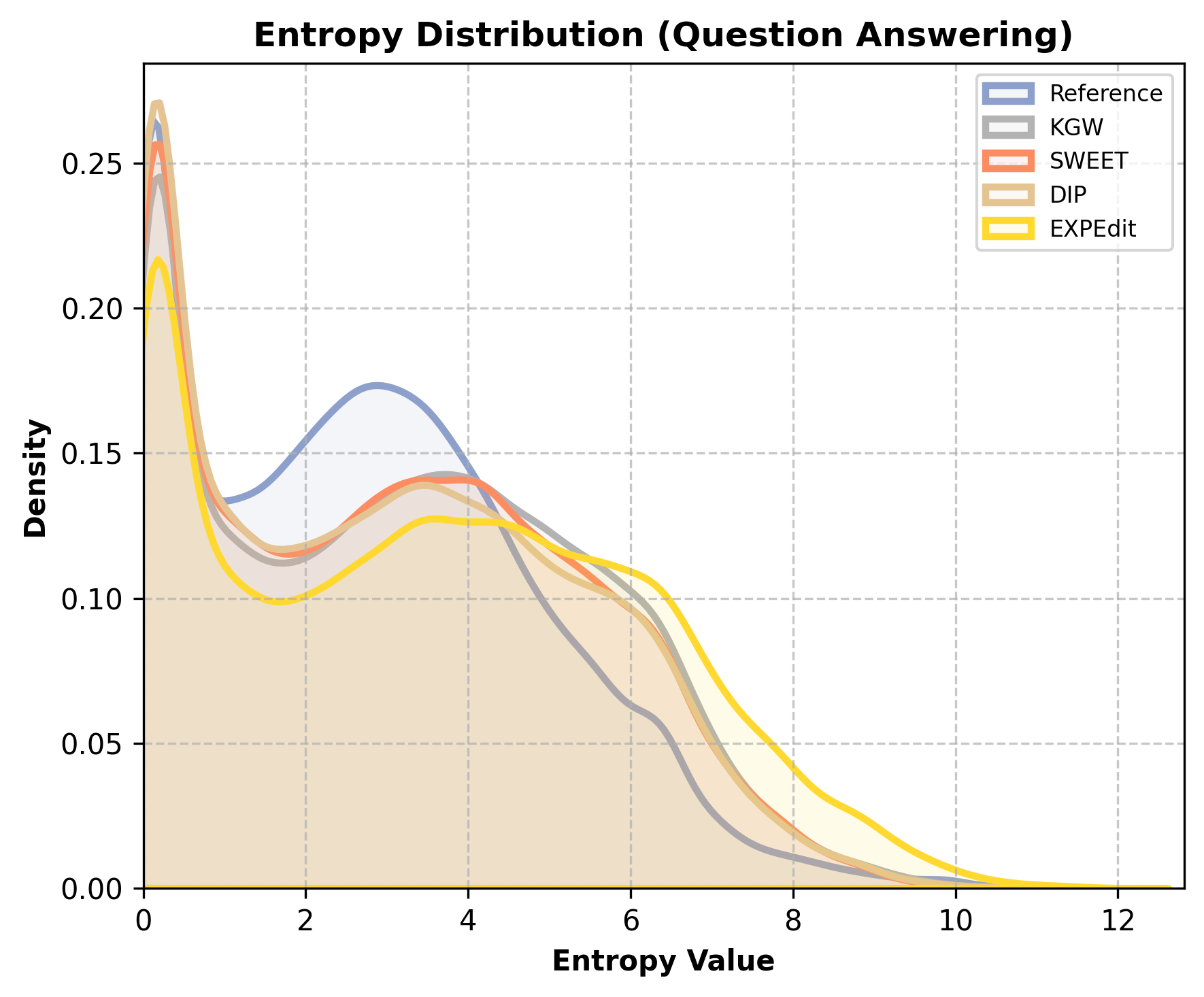}
        \caption{Question Answering}
        \label{fig:subfig1}
    \end{subfigure}
    \hfill
    \begin{subfigure}{0.42\textwidth}
        \centering
        \includegraphics[width=\textwidth]{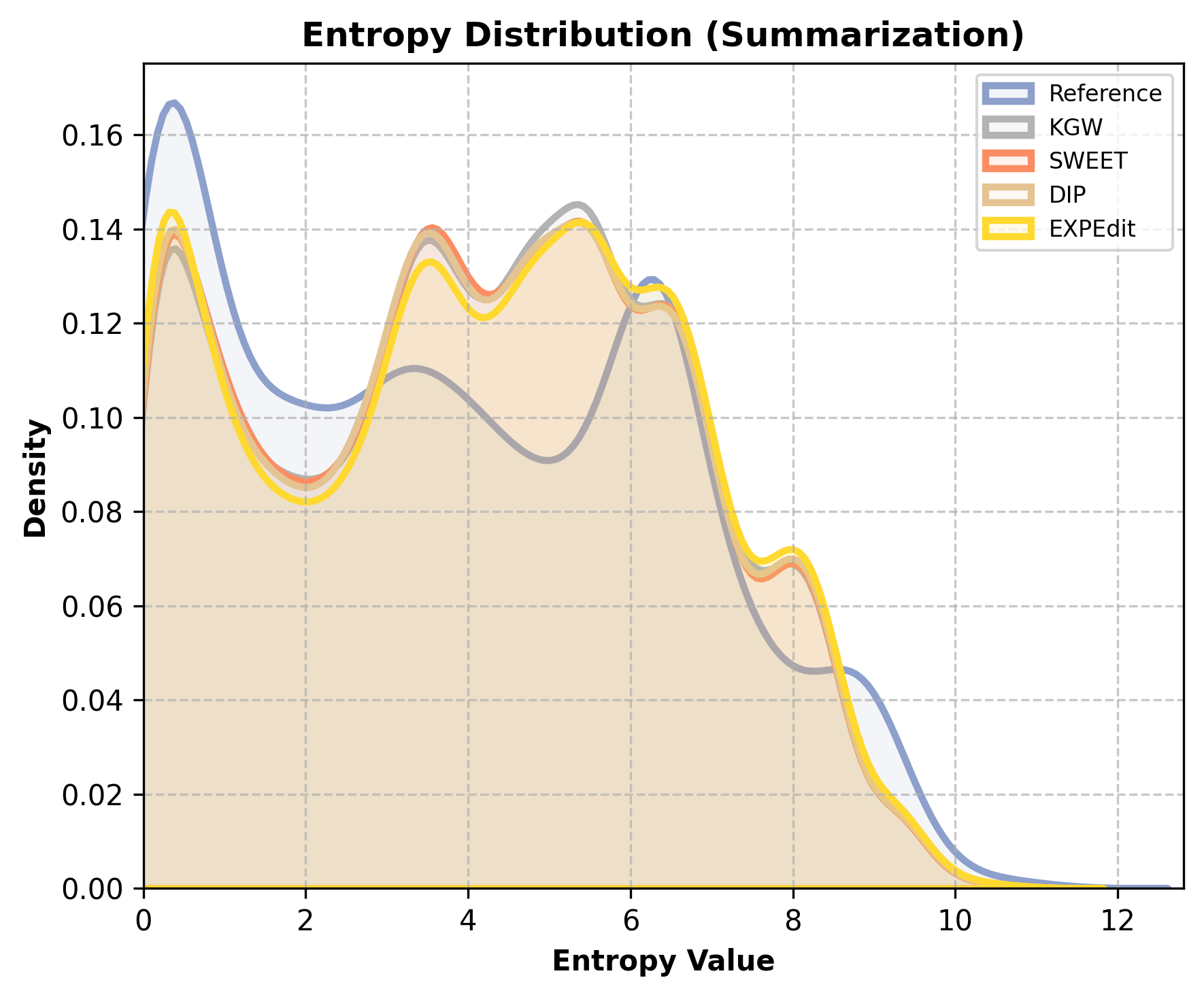}
        \caption{Summarization}
        \label{fig:subfig2}
    \end{subfigure}
    \caption{All watermarking methods cause the entropy distribution to shift in low-to-mid entropy regions for Question Answering and Summarization.}
    \label{fig:app-plot-entropy}
\end{figure}

Token entropy shifts in Question Answering task are more profound in the mid-region entropy as seen in Figure~\ref{fig:subfig1}, where medical terms supposedly exist. This pattern differs in Summarization task where the increase shifts present in the low-entropy region, and significant decrease shifts exist in the mid-entropy region as seen in Figure ~\ref{fig:subfig2}. 

This difference is mainly related to each task having a different text length, QA with 200 tokens and Summarization with fewer than 100 tokens. The QA task, naturally giving the easier prompt for a language model to understand, will generate straightforward answers related to medical topics, thus the general distribution is more preserved than in Summarization task. The summarization, on the other hand, is a question summarization task, which is a more difficult and less straightforward task. The language model has to understand the medical topic and then needs to summarize it according to the given short context limitation, which also potentially becomes the lead of the distorted generated text.

\subsection{Hallucination Samples}
\label{sec:halu}
In Table~\ref{tab:halu}, we showed two samples where the watermarked text introduces hallucination, according to low BERT similarity score. We use a text completion task and pick samples from watermarked text generated using KGW and SWEET. We can see from the table that the extracted entities from the un-watermarked text are still relevant to the original prompt. In contrast, the watermarked texts introduce unrelated terms to the original prompt.

\begin{table}[h]
\small
  \centering
  \begin{tabular}{|p{7.2cm}|}
    \hline
    \multicolumn{1}{|l|}{\textbf{Sample \#1}} \\ \hline
   \textbf{Prompt:} \\ short time . Decongestant tablets or syrup . For example, a medication called pseudoephedrine. This can be obtained from a pharmacy without a prescription. Take the \\ \hline
    \textbf{Un-watermarked Extracted Entities:} \\ pseudoephedrine \\ \hline
    \textbf{KGW-Watermarked Extracted Entities:} \\ pseudoephedrine, \textcolor{red}{amphetamine}, ®, amitriptyline, chlorpheniramine, amylopectin chlorophyllin chlorphenoxyethanone, chlorphenamine, low back pain, sciatica, lower leg numbness, maleate, amiloride, allopurinol/chlorphenaminesine \\ \hline
    \textbf{Hallucination:} One sample from the extracted entities, Amphetamine, is considered below threshold with score -0.204. Amphetamine is not related to decongestant or pseudoephedrine \\
    \hline
    
    \multicolumn{1}{|l|}{\textbf{Sample \#2}} \\ \hline
   \textbf{Prompt:} \\ to develop as a result of long - standing ( chronic ) stress and irritation of a plantar digital nerve . This may be due to the nerve being squashed \\ \hline
    \textbf{Un-watermarked Extracted Entities:} \\
    trauma, haemorrhage, loss of skeletal muscle mass, muscle wasting diseases, obesity, muscle mass, muscle loss \\ \hline
    \textbf{SWEET-Watermarked Extracted Entities:} \\ shoe - counter, fection, infection, chancre, \textcolor{red}{tbc}, herpes zoster, lupus, lupus erythematosus, psoriasis, arthritis, leprosy, verruca, ings \\ \hline
    \textbf{Hallucination:} One sample from the extracted entities, TBC, is considered below threshold with score 0.299. From the prompt, the discussion is about nerve in the foot, which is unrelated to TBC \\
    \hline
  \end{tabular}
  \caption{Hallucination example in Text Generation task.}
  \label{tab:halu}
\end{table}

\section{Implementation Details and Methods Preliminary}
\label{sec:appendix_IMPLEMENTATION}

\subsection{GPT-Judger}
\label{sec:appendix_gptjudge}
The prompt for GPT-Judger formerly referred to the template of \citet{singh2024new}, where the aspects are tailored for the factuality dimension designed in Table~\ref{tab:quality-aspects}. We add detailed questions following each aspect to further trigger deeper reasoning. These questions per aspect are also used as in the Human Evaluation questionnaire detailed in Appendix~\ref{sec:appendix_human}. The example of prompt and response of GPT-Judger for Summarization task is available in Table~\ref{tab:judger-prompt-TC}.

We use OpenAI GPT-4o-2024-08-06 as an LLM evaluator, and in total, we spent around 60 USD on the API usage during experiments.

\begin{table*}
\small
\centering
\caption{GPT-Judger prompt and example response set for Summarization task. We utilize GPT4o-2024-08-06.}
\label{tab:judger-prompt-TC}
\begin{tabular}{p{0.48\textwidth}p{0.48\textwidth}}
\textbf{PROMPT} & \textbf{JUDGER RESPONSE} \\
\hline
Please act as an impartial judge and evaluate the quality of the Summarized Medical Question provided by two large language models to the prompt displayed below. & Overall Superior Response: LLM A provides a more detailed and coherent summarized question while maintaining clarity and relevance to the original query \\
\\
Assess each response according to the criteria outlined, using a 1-5 Likert scale where 1 indicates strong disagreement or the lowest quality, and 5 indicates strong agreement or the highest quality. & \\
\\

Criteria:
\begin{enumerate}
    \item Coherence. Is the language coherence, clear, understandable for a general audience?
    \item Completeness. Does the generated summary miss any important information from the original question?
    \item Factual Accuracy. Does the generated summary introduce any inaccurate or unrelated medical terms not found in the original question?
\end{enumerate}
& LLM A Evaluation:

Coherence: 4 - The language used in LLM A's response is generally clear and understandable. However, the addition of the word "or" between "depression" and "migraines" introduces slight ambiguity, slightly reducing coherence. 

Completeness: 5 - LLM A captures all the critical elements of the original question, such as the focus on vagal nerve stimulation and its potential to alleviate both depression and migraines.

Factual Accuracy: 5 - The summary does not introduce any inaccurate medical terms or unrelated concepts. It stays true to the original topic, discussing the relevance of vagal nerve stimulation to depression and migraines.\\

After scoring each criterion, provide a short summary for each response, including specific examples that influenced your scoring. 
Additionally, don't let the length of the responses influence your evaluation. 
Be as objective as possible and ensure that the order in which the responses are presented does not affect your decision. & LLM B Evaluation: \\
\\
Start with a brief statement about which response you think is superior. Then, for each response and criterion, provide a score, followed by a brief justification for that score. At the very end of your response, declare your verdict by choosing one of the choices below, strictly following the given format: & $\vdots$ \\
\\

[[A]]: [list of scores for LLM A output, in order of Coherence, Completeness, Factual Accuracy] &\\

[[B]]: [list of scores for LLM B output, in order of Coherence, Completeness, Factual Accuracy] &\\ 
\\

[Prompt] & Verdict \\

[LLM A's Answer]& [[A]]: [4, 5, 5]\\

[LLM B's Answer] & [[B]]: [3, 3, 4]\\      

\hline
\end{tabular}
\label{tab:gpt-judger-prompt}
\end{table*}

\subsection{Human Evaluation}
\label{sec:appendix_human}
\begin{figure*}
    \centering
    \begin{subfigure}[b]{0.6\textwidth}
        \adjustbox{frame, max width=\textwidth}{%
        \includegraphics[width=\textwidth]{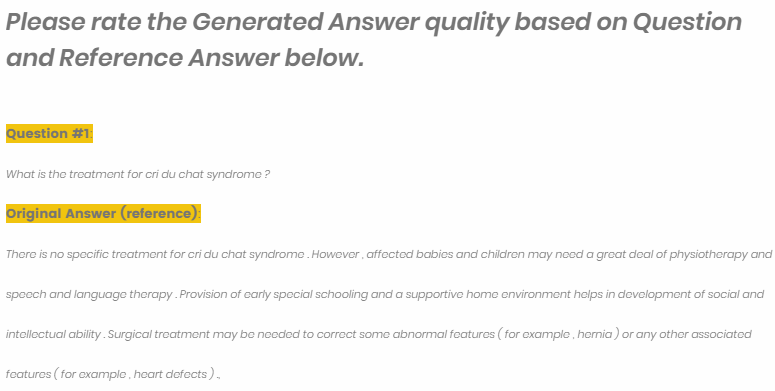}%
    }
        \caption{Reference answer help non-expert respondent evaluate the watermarked text.}
    \end{subfigure}
    \hfill
    \begin{subfigure}[b]{0.6\textwidth}
        \adjustbox{frame, max width=\textwidth}{%
        \includegraphics[width=\textwidth]{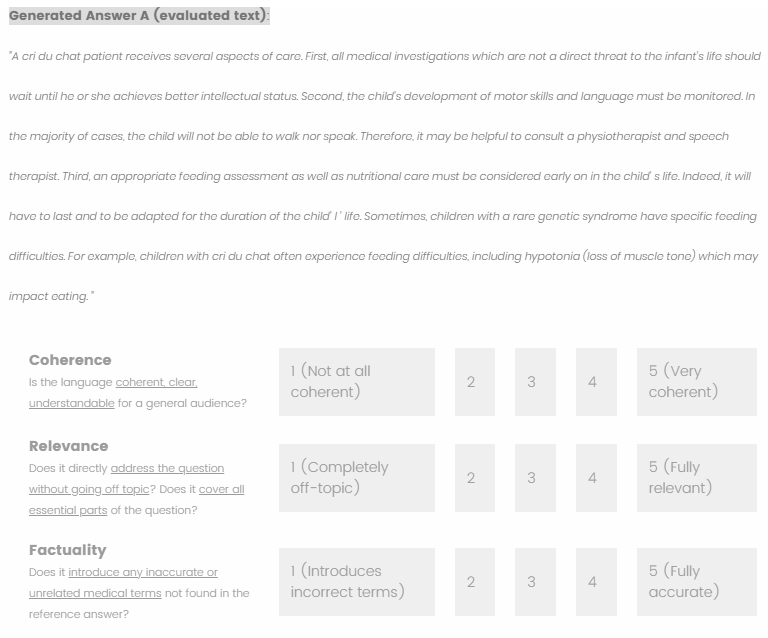}%
    }
        \caption{Generated Answer A: watermarked text or un-watermarked text.}
    \end{subfigure}
     \hfill
    \begin{subfigure}[b]{0.6\textwidth}
        \adjustbox{frame, max width=\textwidth}{%
        \includegraphics[width=\textwidth]{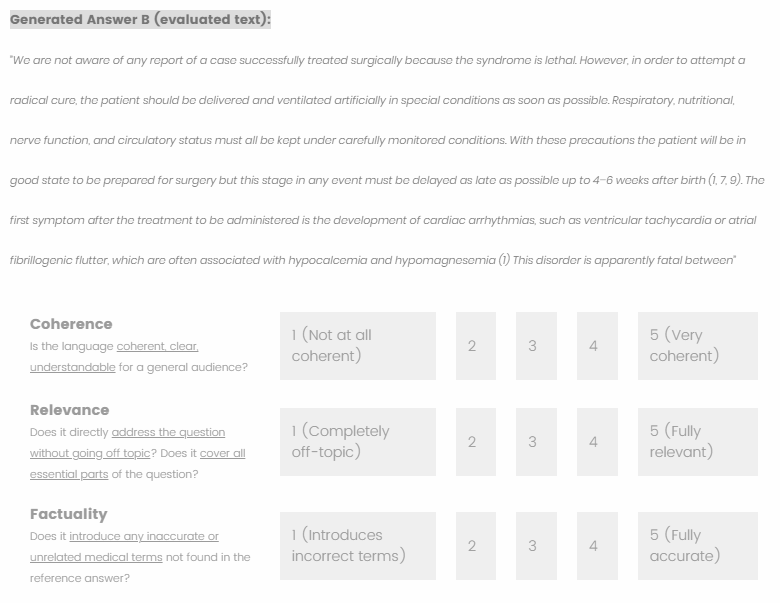}%
    }
        \caption{Generated Answer B: watermarked text or un-watermarked text.}
    \end{subfigure}
    \caption{Questionnaire for human evaluation on Question Answering task. Respondents were presented Ground-truth text from the dataset, Watermarked text, and Un-watermarked text at the same page to fairly evaluate each generated text. Positioning for the generated text was randomized to prevent bias. }
    \label{fig:questionnaire}
\end{figure*}

The human evaluation is conducted for the QA task as the nature of the task is typically easier for humans to interpret and rate based on the ground-truth answer, compared to other tasks like Text Completion. Sample of questionnaire shown in Figure~\ref{fig:questionnaire}. We collected results from 6 respondents (4 graduate students and 2 medical practitioners) to evaluate sample of 10\% of the total generated texts, while making sure each item is rated by 3 different respondents. This study has been approved by IRB under exempt human subject research with IRB ID STUDY00007853. Respondents were recruited through social media groups and domain background were checked later. All participation was voluntary and was clearly stated in the consent page of the questionnaire.

\subsection{Tasks Details}
\label{sec:appendix_tasks}

\subsubsection{Text Completion}
For this task, we utilize the HealthQA dataset \cite{zhu2019hierarchical}, which consists of medical question-answer pairs. We focus solely on the answer portions, selecting 200 answers that exceed 230 words in length. From each answer, we extract the last 230 words, using the first 30 words as the input prompt and tasking the model with generating the subsequent 200 words. This experimental design aligns with prior watermarking evaluation frameworks \cite{kirchenbauer2023watermark}, allowing for consistent comparison across domains.

\subsubsection{Question Answering}
We employ the same HealthQA dataset \cite{zhu2019hierarchical} for our QA task, but utilize both question and answer components. To maintain consistency with the Text Completion task, we select 200 data points containing questions of exactly 10 words and answers of fewer than 250 words. The questions (ending with \texttt{?}) serve as input prompts without additional instructions, while the corresponding answers constitute the ground truth for evaluation.

\subsubsection{Summarization}
For the Summarization task, we employ the MeQSum dataset \cite{abacha2019summarization}, which provides pairs of detailed clinical questions and their concise summaries. We construct model prompts by prefixing the clinical question text with \texttt{Write a short question that summarizes this question:} and appending \texttt{Summarized Question:} after the question text. To introduce diversity, we limit our selection to inputs with a maximum of 60 words and summaries containing at least 10 words, creating more variation in input-output length ratios compared to the other tasks.

\subsection{Watermarking Methods}
\label{sec:appendix_method}

\subsubsection{KGW \cite{kirchenbauer2023watermark}}
We provide brief preliminaries for KGW watermarking approach. For a given language model $f_{\text{LM}}$ with vocabulary $\mathcal{V}$, the likelihood probability of a token $y_t$ is calculated as follows:
\begin{align}
l_t &= f_{\text{LM}}(x, y_{[:t]}) \\
p_{t,i} &= \frac{e^{l_{t,i}}}{\sum_{j=1}^{|\mathcal{V}|} e^{l_{t,j}}}
\end{align}
\noindent where $x = {x_0, \ldots, x_{M-1}}$ and $y_{[:t]} = {y_1, \ldots, y_{t-1}}$ are an $M$-length tokenized prompt and the generated token sequence respectively, and $l_t \in \mathbb{R}^{|\mathcal{V}|}$ is the logit vector.

\noindent\textbf{Watermarking}\hspace{0.5em} 
In the watermarking procedure, the entire tokens in $\mathcal{V}$ at each time-step are randomly binned into green ($\mathcal{G}_t$) and red ($\mathcal{R}_t$) groups in proportions of $\gamma$ and $1 - \gamma$ ($\gamma \in (0, 1)$), respectively. The method increases the logits of green group tokens by adding a fixed bias $\delta$, promoting these tokens to be sampled at each position. Consequently, watermarked LM-generated text is more likely than $\gamma$ to contain tokens from the green group. In contrast, since humans have no knowledge of the hidden green-red partition rule, the proportion of green group tokens in human-written text is expected to be close to $\gamma$.

\noindent\textbf{Detection}\hspace{0.5em} 
The presence of watermarking in text is detected through a one-sided $z$-test by testing the null hypothesis that the text is not watermarked. The $z$-score is calculated using the number of recognized green tokens in the test text. Subsequently, the text is determined to be watermarked if the $z$-score exceeds a predetermined threshold $z_{\text{threshold}}$.
\subsubsection{SWEET \cite{lee-etal-2024-wrote}}
SWEET improve KGW \cite{kirchenbauer2023watermark} by distinguishing watermark applicable tokens, meaning it embed and detect watermarks only within tokens with high entropy. 

\noindent\textbf{Watermarking}\hspace{0.5em} 
Given a tokenized prompt $x = {x_0, \ldots, x_{M-1}}$ and already generated tokens $y_{[:t]} = {y_0, \ldots, y_{t-1}}$, a model calculates an entropy value ($H_t$) of the probability distribution for $y_t$.The watermarking only applied when $H_t$ exceeds the threshold, $\tau$. Then, randomly split the vocabulary into green and red groups with a fixed green token ratio $\gamma$. If a token is selected to be watermarked, a constant $\delta$ is added to the logits of green tokens, thereby promoting the sampling of these tokens. By limiting the promotion of green tokens only to positions with high entropy, it prevent modifications to the model's logit distribution for tokens where the model has high confidence (and, therefore, low entropy).

\noindent\textbf{Detection}\hspace{0.5em} 
Given a token sequence $y = {y_0, \ldots, y_{N-1}}$, the task is to detect watermarks within $y$, thereby determining whether it was generated by the specific language model. As in the generation phase, SWEET computes the entropy values $H_t$ for each $y_t$. Let $N_h$ denote the number of tokens that have an entropy value $H_t$ higher than the threshold $\tau$, and let $N_h^G$ denote the number of green tokens among those in $N_h$. Finally, with the green list ratio $\gamma$ used in the generation step, SWEET computes a z-score under the null hypothesis that the text is not watermarked:
\begin{equation}
z = \frac{N_h^G - \gamma N_h}{\sqrt{N_h \gamma (1-\gamma)}}
\end{equation}
The presence of a watermark can be assessed with increasing confidence as the z-score increases. SWEET sets $z_{\text{threshold}}$ as a cut-off score. If $z > z_{\text{threshold}}$ holds, it determines that the watermark is embedded in $y$ and thus the text was generated by the LLM. 

\subsubsection{EXP-edit \cite{kuditipudi2024robust}}
This method apply the watermarking via Exponential Minimum Sampling \cite{aaronson2022watermarking} with slight modification of detection using Levenshtein Alignment.

\noindent\textbf{Watermarking}\hspace{0.5em} 
Given a tokenized prompt $x = {x_0, \ldots, x_{M-1}}$ and already generated tokens $y_{[:t]} = {y_0, \ldots, y_{t-1}}$, a watermark key sequence $\xi \in [0, 1]^N$ is used to deterministically map to samples from the language model. The decoder function $\Gamma$ for each token is defined as:
\begin{equation}
\Gamma(\xi, \mu) := \arg\min_{i \in [N]} -\log(\xi_i)/\mu(i)
\end{equation}
\noindent where $\mu \in \Delta([N])$ is the probability distribution over the vocabulary from the language model. This decoder is distortion-free, as marginalizing over the watermark key sequence, the distribution of generated tokens remains equivalent to sampling directly from the language model.

\noindent\textbf{Detection}\hspace{0.5em} 
Given a token sequence $y = {y_0, \ldots, y_{N-1}}$ that might be watermarked, a Levenshtein alignment cost function $d_\gamma$ allows for robust detection even when the text has been modified through insertions or deletions:
\begin{equation}
d_\gamma(y, \xi) := \min \begin{cases}
d_\gamma(y_{2:}, \xi_{2:}) + \\ 
d_0(y_1, \xi_1) \
d_\gamma(y, \xi_{2:}) + \\ \gamma \
d_\gamma(y_{2:}, \xi) + \\ \gamma
\end{cases}
\end{equation}

\noindent where $d_0(y, \xi) = \sum_{i=1}^{\text{len}(y)} \log(1 - \xi_{i,y_i})$ is the base alignment cost for EXP, and $\gamma$ is a hyperparameter controlling the penalty for insertions and deletions. For EXP-edit, the optimal value is $\gamma = 0.0$.
The detection algorithm computes a test statistic by finding the minimum cost alignment between blocks of the input text and the watermark key sequence. A permutation test is then used to generate a p-value by comparing this test statistic against randomly generated watermark keys, determining whether the text is likely watermarked.

\subsubsection{DiPmark \cite{wu2024resilient}}
DiPmark operates on a vocabulary set $V$ with size $N = |V|$ and employs a distribution-preserving reweight strategy that modifies token probabilities without changing the original distribution $P_M(x_{n+1} | x_{1:n})$.

\noindent\textbf{Watermarking}\hspace{0.5em} 
During generation, the system derives a texture key $s_i$ from previously generated tokens and combines it with a secret key $k$ to produce a cipher $\theta_i = h(k, s_i)$ using a hash function $h$. This cipher represents a permutation of the vocabulary tokens. The core of DiPmark is its DiP-reweight strategy, defined as:
\begin{equation}
\begin{aligned}[t]
&P_W(t_i|x, \theta) := \\
&(1 - \alpha)P_W^\alpha(t_i|x, \theta) + \alpha P_W^{1-\alpha}(t_i|x, \theta)
\end{aligned}
\end{equation}
where $\alpha \in [0,1]$ is the reweight parameter and $P_W^\alpha$ adjusts token probabilities within interval $[0, \alpha]$ to 0 and scales others by $\frac{1}{1-\alpha}$. The next token is then sampled according to this modified distribution. This approach mathematically guarantees that $E_\theta[P_W(t|x, \theta)] = P_M(t|x)$, ensuring the original distribution is preserved across multiple generations, which distinguishes DiPmark from other watermarking approaches.

\noindent\textbf{Detection}\hspace{0.5em} 
The detection process uses a statistical test based on the "green token ratio" to identify watermarked text. For a given text sequence $x_{1:n}$, the detector computes the cipher $\theta_i$ for each position using the same hash function and secret key. Each permutation $\theta_i$ is divided using a green list separator $\gamma \in [0,1]$, with the last $(1-\gamma)N$ tokens in the permutation designated as "green tokens." 

The detector counts the number of green tokens $L_G(\gamma)$ in the sequence and calculates the green token ratio:
\begin{equation}
\Phi(\gamma, x_{1:n}) := \frac{L_G(\gamma)}{n} - (1 - \gamma)
\end{equation}

This ratio is compared against a threshold $z$; if $\Phi > z$, the text is classified as watermarked. 


\end{document}